%% file: main.tex
\documentclass[11pt,a4paper]{article}
\usepackage{times,latexsym}
\usepackage{url}
\usepackage{adjustbox}
\usepackage[T1]{fontenc}
\usepackage{graphicx}
\usepackage{amssymb}
\usepackage{subcaption}
\usepackage{xcolor}
\usepackage{float}
\usepackage{booktabs}
\usepackage[normalem]{ulem}

\definecolor{my_yellow}{RGB}{255, 242, 204}
\definecolor{my_blue}{RGB}{211, 215, 230}

\usepackage[acceptedWithA]{tacl2021v1}

\usepackage{xspace,mfirstuc,tabulary}

\newcommand{\mightomit}[1]{{\color{red}\sout{}}}

\newif\iftaclinstructions
\taclinstructionsfalse 
\iftaclinstructions
\renewcommand{\confidential}{}
\renewcommand{\anonsubtext}{(No author info supplied here, for consistency with
TACL-submission anonymization requirements)}
\newcommand{\instr}
\fi

\iftaclpubformat 

\else

\fi

\title{PADA: Example-based Prompt Learning \\ for on-the-fly Adaptation to Unseen Domains}

\author{Eyal Ben-David \Thanks{Both authors equally contributed to this work.}\\
\And Nadav Oved \footnotemark[\value{footnote}] \\
 Technion - Israel Institute of Technology\\
{\tt \{eyalbd12@campus.$|$nadavo@campus.$|$roiri@\}technion.ac.il}
\And Roi Reichart\\
}

\date{}

\begin{document}
\maketitle

\input{0.abstract}

\input{1.intro}

\input{2.related}

\input{3.problem}
\input{4.model}

\input{5.experimental-setup}
\input{6.results}

\input{7.ablation}

\input{8.discussion}

\section*{Acknowledgements}
We would like to thank the action editor and the reviewers, as well as the members of the IE@Technion NLP group for their valuable feedback and advice. This research was partially funded by an ISF personal grant No. 1625/18.

\bibliography{tacl2021}
\bibliographystyle{acl_natbib}

\end{document}

%% file: 0.abstract.tex
\begin{abstract}
Natural Language Processing algorithms have made incredible progress, but they still struggle when applied to out-of-distribution examples. We address a challenging and underexplored version of this domain adaptation problem, where an algorithm is trained on several source domains, and then applied to examples from unseen domains that are unknown at training time. Particularly, no examples, labeled or unlabeled, or any other knowledge about the target domain are available to the algorithm at training time. We present \emph{PADA}: An example-based autoregressive Prompt learning algorithm for on-the-fly Any-Domain Adaptation, based on the T5 language model. Given a test example, \emph{PADA} first generates a unique prompt for it and then, conditioned on this prompt, labels the example with respect to the NLP prediction task. PADA is trained to generate a prompt which is a token sequence of unrestricted length, consisting of Domain Related Features (DRFs) that characterize each of the source domains. Intuitively, the generated prompt is a unique signature that maps the test example to a semantic space spanned by the source domains. In experiments with 3 tasks (text classification and sequence tagging), for a total of 14 multi-source adaptation scenarios, \emph{PADA} substantially outperforms strong baselines.\footnote{Our code and data are available at \url{https://github.com/eyalbd2/PADA}.}\footnote{An earlier version of this paper was previously uploaded to the arxiv under the name: "PADA: A Prompt-based Autoregressive Approach for Adaptation to Unseen Domains"}
\end{abstract}

%% file: 1.intro.tex
\begin{figure}[ht]
\includegraphics[scale=0.53]{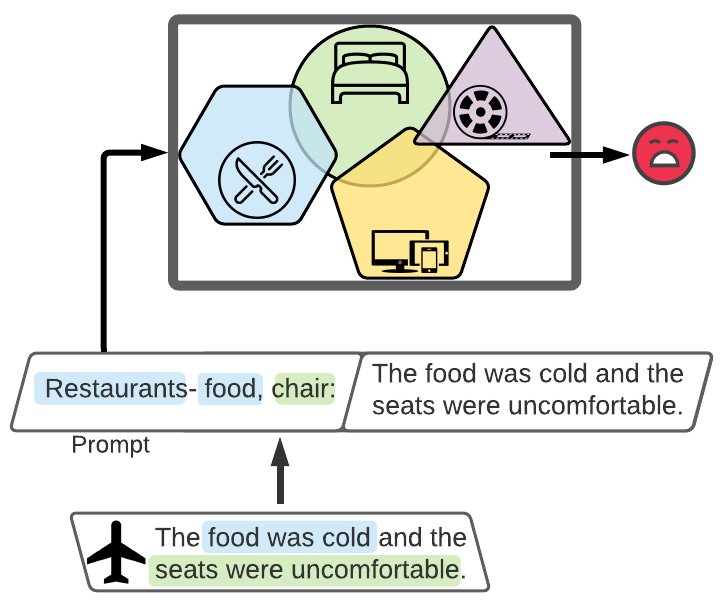}
\centering
\caption{Text classification with PADA. Colored texts signify relation to a specific source domain. \emph{PADA} first generates the domain name, followed by a set of DRFs related to the input example. Then it uses the prompt to predict the task label.}
\centering
\label{fig:any-domain}
\end{figure}

\section{Introduction}
\label{sec:intro}

Natural Language Processing (NLP) algorithms are gradually achieving remarkable milestones \citep{DBLP:conf/naacl/DevlinCLT19, lewis-etal-2020-bart, DBLP:conf/nips/BrownMRSKDNSSAA20}. However, such algorithms often rely on the seminal assumption that the training set and the test set come from the same underlying distribution. Unfortunately, this assumption often does not hold since text may emanate from many different sources, each with unique distributional properties. As generalization beyond the training distribution is still a fundamental challenge, NLP algorithms suffer a significant degradation when applied to out-of-distribution examples.

Domain Adaptation (DA) explicitly addresses the above challenge, striving to improve out-of-distribution generalization of NLP algorithms. DA algorithms are trained on annotated data from source domains, to be effectively applied in a variety of target domains.  Over the years, considerable efforts have been devoted to the DA challenge, focusing on various scenarios where the target domain is known at training time (e.g. through labeled or unlabeled data) but is yet under-represented \citep{roark2003supervised, daume2006domain, reichart2007self, mcclosky2010automatic, rush2012improved, schnabel2014flors}. 
Still, the challenge of adaptation to \textit{any possible target domain}, that is unknown at training time, is underexplored in DA literature.\footnote{The any-domain adaptation setting is addressed in the model robustness literature. In \S\ref{sec:any-domain}, we discuss the differences between these static methods and our dynamic approach.}

In this work, we focus on adaptation to any target domain, which we consider a ``Holy Grail'' of DA (\S\ref{sec:any-domain}). Apart from the pronounced intellectual challenge, it also presents unique modeling advantages as target-aware algorithms typically require training a separate model for each target domain, leading to an inefficient overall solution. 


Intuitively, better generalization to unseen domains can be achieved by integrating knowledge from several source domains. We present \emph{PADA}: An example-based autoregressive Prompt learning algorithm for on-the-fly Any-Domain Adaptation (\S\ref{sec:prompt-based-da}), which utilizes an autoregressive language model (T5, \citet{raffel2020exploring}), and presents a novel mechanism which learns to generate human-readable prompts that represent multiple source domains. Given a new example, from any unknown domain, the model first generates properties (a sequence of tokens) that belong to familiar (source) domains and relate to the given example. Then, the generated sequence is used as a prompt for the example, while the model performs the downstream task.\footnote{We use a language model, pre-trained on massive unlabeled data, and it is possible that this model was exposed to text from the source or target domains. Yet, the downstream task training is based only on examples from the source domains without any knowledge of future target domains.}
\emph{PADA} implements a specialized two-stage multi-task protocol which facilitates model parameter sharing between the prompt generation and the downstream tasks. 
Ultimately, \emph{PADA} performs its adaptation per example, by leveraging (1) an example-specific prompting mechanism and (2) a two-stage multi-task objective.

In order to generate effective prompts, we draw inspiration from previous work on pivot features \citep{DBLP:conf/emnlp/BlitzerMP06, ziser2018pivot, ben2020perl} to define sets of Domain Related Features (DRFs, \S\ref{sec:DRF}). DRFs are tokens which are strongly associated with one of the source domains, encoding domain-specific semantics. We leverage the DRFs of the various source domains in order to span their shared semantic space. Together, these DRFs reflect the similarities and differences between the source domains, in addition to domain-specific knowledge.

Consider the task of review sentiment classification (Figure~\ref{fig:any-domain}). The model is familiar with four source domains: \textit{restaurants}, \textit{home-furniture}, \textit{electronic-devices}, and \textit{movies}. When the model encounters a review, this time from the \textit{airlines} domain, it uses DRFs to project the example into the shared semantic space, via the prompting mechanism. In the given example the DRFs marked in blue and green relate to the \textit{restaurants} and the \textit{home-furniture} domains, respectively. The DRF-based prompt is then used in classification.

We evaluate \emph{PADA} in the multi-source DA setting, where the target domain is unknown during training (\S \ref{sec:experiments}, \ref{sec:results}).  We consider two text classification tasks (Rumour Detection and  Multi-Genre Natural Language Inference (MNLI)), and a sequence tagging task (Aspect Prediction), for a total of $14$ DA setups. \emph{PADA} outperforms  strong baselines, yielding substantial error reductions. 


%% file: 2.related.tex
\section{Related Work}
\label{sec:related}

We first describe research in the setting of unsupervised DA with a focus on pivot-based methods. We then continue with the study of DA methods with multiple sources, focusing on mixture of experts models. Finally, we describe autoregressive language models and prompting mechanisms, and the unique manner in which we employ T5 for DA.


\paragraph{Unsupervised Domain Adaptation (UDA)}

With the breakthrough of deep neural network (DNN) modeling, attention from the DA community has been directed to representation learning approaches. 
One line of work employs DNN-based autoencoders to learn latent representations. These models are trained on unlabeled source and target data with an input reconstruction loss \citep{DBLP:conf/icml/GlorotBB11, DBLP:conf/icml/ChenXWS12, DBLP:conf/acl/YangE14, DBLP:journals/jmlr/GaninUAGLLML16}. Another branch employs pivot features to bridge the gap between a source domain and a target domain \citep{DBLP:conf/emnlp/BlitzerMP06, DBLP:conf/acl/BlitzerDP07, DBLP:conf/www/PanNSYC10}. Pivot features are prominent to the task of interest and are abundant in the source and target domains. Recently, \citet{DBLP:conf/conll/ZiserR17, ziser2018pivot, DBLP:conf/acl/ZiserR19}  married the two approaches. Later on, \citet{DBLP:conf/emnlp/HanE19} presented a pre-training method, followed by \citet{ben2020perl} and \citet{lekhtman2021dilbert} who introduced a pivot-based variant for pre-training contextual word embeddings.


Crucially, UDA models assume access to unlabeled data from the target domain in-hand during training. We see this as a slight relaxation to the goal of generalization beyond the training distribution. Moreover, this definition has engineering disadvantages, as a new model is required for each target domain. To this end, we pursue the any-domain adaptation setting, where unlabeled target data is unavailable at training time.

We draw inspiration from pivot-based modeling. The pivot definition relies on labeled source domain data and unlabeled source and target domain data (which is unavailable in our setup). Particularly, good pivots are ones that are correlated with the task label. 
Hence, pivot features are typically applied to tasks which offer meaningful correlations between words and the task label, such as sentiment classification. For other types of tasks, pivots may be difficult to apply. Consider the \textit{MNLI} dataset, where the task is to understand the directional relation between a pair of sentences (entailment, contradiction or neutral). In such a task it is unlikely to find meaningful correlations between single words and the label.
Instead, we define task-invariant DRFs, features which are highly correlated with the identity of the domain. Since domains are highly correlated with words, our DRFs are lexical in nature.

Our proposed approach is an important step forward from pivots, as our model generates DRF sequences of unrestricted lengths, instead of focusing on individual words. 
Moreover, pivots are typically applied in single source setups, and while our method can operate with a single source domain, we utilize multiple source domains to facilitate generalization to unknown target domains.


\paragraph{Multi-Source Domain Adaptation}
Most existing \textit{multi-source DA} methods follow the setup definitions of unsupervised DA, while considering more than one source domain. A prominent approach is to fuse models from several sources. Early work trained a classifier for each domain and assumed all source domains are equally important for a test example \citep{DBLP:conf/acl/LiZ08, DBLP:conf/cikm/LuoZHXH08}. More recently, adversarial-based methods used unlabeled data to align the source domains to the target domains \citep{DBLP:conf/iclr/0002ZWCMG18, DBLP:conf/naacl/ChenC18}. Meanwhile, \citet{DBLP:conf/acl/KimSK17} and \citet{DBLP:conf/emnlp/GuoSB18} explicitly weighted a Mixture of Experts (MoE) model based on the relationship between a target example and each source domain. However, \citet{wright2020transformer} followed this work and tested a variety of weighting approaches on a Transfomers-based MoE and found a naive weighting approach to be very effective.

We recognize two limitations in the proposed MoE solution. First, MoE requires training a standalone expert model for each source domain. Hence, the total number of parameters increases (typically linearly) with the number of source domains, which harms the solution's scalability.
One possible solution could be to train smaller-scale experts \citep{DBLP:conf/emnlp/PfeifferVGR20, DBLP:conf/emnlp/RucklePG20}, but this approach is likely to lead to degradation in performance. 
Second, domain experts are tuned towards domain-specific knowledge, at times at the expense of cross-domain knowledge which highlights the relationship between different domains. In practice, test examples may arrive from unknown domains, and may reflect a complicated combination of the sources. To cope with this, MoE ensembles the predictions of the experts using heuristic methods, such as a simple average or a weighted average based on the predictions of a domain-classifier. Our results indicate that this approach is sub-optimal.

Moreover, we view domain partitioning as often somewhat arbitrary (consider for example the differences between the \textit{dvd} and \textit{movie} domains). We do not want to strictly confine our model to a specific partitioning and rather encourage a more lenient approach towards domain boundaries. Hence, in this work, we train only a single model which shares its parameters across all domains.
Furthermore, we are interested in adapting to any target domain, such that no information about potential target domains is known at training time. Some of the above works \citep{wright2020transformer} in fact avoid utilizing target data, thus they fit the any-domain setting and form two of our baselines. Yet, in contrast to these works, the \textit{any-domain} objective is a core principle of this study.

\paragraph{Autoregressive LMs and Prompting}
Recently, a novel approach to language modeling has been proposed, which casts it as a sequence-to-sequence task, by training a full Transformer (encoder-decoder) model \citep{DBLP:conf/nips/VaswaniSPUJGKP17} to autoregressively generate masked, missing or perturbed token spans from the input sequence \citep{raffel2020exploring, lewis-etal-2020-bart}. 
\citet{raffel2020exploring} present a particularly interesting approach with the T5 model. It treats all tasks as generative (text-to-text), eliminating the need for a task-specific network architecture. This is made possible by prefixing each example with a prompt phrase denoting the specific task being performed.

Recent works have further explored such prompting mechanisms in several avenues: Adapting a language model for different purposes \citep{DBLP:conf/nips/BrownMRSKDNSSAA20}; eliciting sentiment or topic-related information \citep{DBLP:journals/tacl/JiangXAN20,Sun2020ConditionedNL,DBLP:conf/emnlp/ShinRLWS20,DBLP:conf/eacl/HavivBG21}; efficient fine-tuning \citep{DBLP:conf/acl/LiL20,DBLP:conf/naacl/ScaoR21}; or as a method for few-shot learning \cite{DBLP:conf/acl/GaoFC20,DBLP:conf/eacl/SchickS21}.\footnote{For a comprehensive discussion of the research on prompting mechanisms, we refer to \citet{DBLP:journals/corr/abs-2107-13586}.}
In this work, we make use of T5's prompting mechanism as a way of priming the model to encode domain-specific characteristics relating to each example from an unknown target domain. Borrowing terminology from \citet{DBLP:journals/corr/abs-2107-13586}, our approach falls under the ``Prompt+LM Tuning'' training strategy \citep{DBLP:journals/corr/abs-2103-10385, DBLP:journals/corr/abs-2105-11259}. In this strategy, prompt-relevant parameters are fine-tuned together with some or all of the parameters of the pre-trained model (T5 in our case). However, in contrast to prompt tuning approaches which focus on representation level tuning \citep{DBLP:journals/corr/abs-2103-10385, DBLP:conf/acl/LiL20, DBLP:journals/corr/abs-2104-08691}, we train T5 to generate human readable prompts consisting of natural language tokens that encode domain-specific information relating to the the given example. To the best of our knowledge, this work is the first to learn to generate textual prompts alongside a downstream prediction task. It is also the first to generate a unique prompt per example. Finally, it is the first to design a prompting mechanism for the purpose of DA.




%% file: 3.problem.tex
\section{Any-Domain Adaptation}
\label{sec:any-domain}


\paragraph{DA and Transfer Learning}
A prediction task (e.g., Rumour Detection) is defined as $\mathcal{T}=\{\mathcal{Y}\}$, where $\mathcal{Y}$ is the task's label space. 
We denote $\mathcal{X}$ to be a feature space, $P(X)$ to be the marginal distribution over $\mathcal{X}$, and $P(Y)$ the prior distribution over $\mathcal{Y}$. The domain is then defined by $\mathcal{D}^\mathcal{T}=\{\mathcal{X},P(X), P(Y), P(Y|X)\}$. DA is a particular case of transfer learning, namely \textit{transductive transfer learning} \citep{DBLP:conf/coling/RamponiP20}, in which $\mathcal{T}_S$ and $\mathcal{T}_T$, the source and target tasks, 
are the same. However, $\mathcal{D}^\mathcal{T}_S$ and $\mathcal{D}^\mathcal{T}_T$, the source and target domains, differ in at least one of their underlying probability distributions, $P(X)$, $P(Y)$, or $P(Y|X)$.\footnote{In  \textit{inductive transfer learning} $\mathcal{T}_S$ differs from $\mathcal{T}_T$.} The goal in DA is to learn a function $f$ from a set of source domains  $\{\mathcal{D}_{S_i}\}_{i=1}^{K}$ that generalizes well to a set of target domains $\{\mathcal{D}_{T_i}\}_{i=1}^{M}$.

\paragraph{The Any-Domain Setting}
We focus on building an algorithm for a given task, that is able to adapt to \textit{any-domain}. To this end, we assume zero knowledge about the target domain, $\mathcal{D}_T$, at training time. Hence, we slightly modify the classic setting of unsupervised multi-source domain adaptation, by assuming we have no knowledge or access to labeled or unlabeled data from the target domains. We only assume access to labeled training data from $K$ source domains $\{\mathcal{D}_{S_i}\}_{i=1}^{K}$, where $\mathcal{D}_{S_i} \triangleq \{ (x_{t}^{S_i}, y_{t}^{S_i}) \}_{t=1}^{n_i}$. The goal is to learn a model using only the source domains data, that generalizes well to unknown target domains.

The NLP and ML literature addresses several settings that are similar to any-domain adaptation. However, our on-the-fly example-based approach is novel. Below, we discuss these settings and the differences between their proposed solution approaches and ours.

The goal of any-domain adaptation was previously explored through the notion of \textit{domain robustness}. Algorithms from this line of work seek generalization to unknown distributions through optimization methods which favor robustness over specification \cite{DBLP:conf/icml/HuNSS18, DBLP:conf/emnlp/OrenSHL19, DBLP:conf/iclr/SagawaKHL20, DBLP:journals/corr/abs-2012-07421, wald2021on}. This is typically achieved by training the model to focus on domain-invariant features, which are considered fundamental to the task and general across domains \cite{DBLP:conf/icml/MuandetBS13, DBLP:journals/jmlr/GaninUAGLLML16, DBLP:journals/corr/abs-1907-02893, DBLP:conf/amta/MullerRS20}. In contrast, this work proposes to achieve this goal through on-the-fly example-based adaptation, utilizing both domain-invariant and domain-specific features, as the latter often proves relevant to the new domain \cite{DBLP:conf/emnlp/BlitzerMP06, DBLP:conf/conll/ZiserR17}. For instance, consider the example presented in Figure~\ref{fig:any-domain}. The expression ``food was cold'' would be considered as domain-specific, considering the \textit{restaurants} domain. Despite it not being a domain-invariant feature, it may serve as a valuable feature for the target domain (\textit{airlines}).


\textit{Any-domain adaptation} also draws some similarities with the \textit{continual learning} \citep{DBLP:phd/dnb/Ring95} and \textit{zero-shot learning} \citep{DBLP:conf/nips/PalatucciPHM09} paradigms. \textit{Continual learning} systems seek to transfer knowledge from a number of known tasks to a new one, while in our proposed setting new domains arrive during inference, and as opposed to continual learning, we do not update the parameters of the model when a new domain is presented (we actually do not even know the domains of the test examples).\footnote{\citet{DBLP:conf/iclr/OswaldHSG20} explore the notion of inferring the new example's task out of the training tasks.}
The \textit{zero-shot} setting also does not update the parameters of the model given a new task, yet its definition is less consistent across different models: GPT-3 \citep{DBLP:conf/nips/BrownMRSKDNSSAA20} attempts to transfer knowledge to an unknown target task $\mathcal{T}_T$ and unknown domain $\mathcal{D}_T$;
\citet{blitzer2009zero} assume access to unlabeled data from various domains including the target domain; and \citet{DBLP:conf/eccv/PengWE18} use data of a different task from the target domain. 
In contrast, our problem setting specifically focuses on domain adaptation, while assuming no prior knowledge of the target domain.


The any-domain adaptation setting naturally calls for an example-level adaptation approach. Since the model does not have any knowledge about the target domain during training, each example it encounters during inference should be aligned with the source domains.

%% file: 4.model.tex
\section{Example-based Adaptation through Prompt Learning}
\label{sec:prompt-based-da}

In this work 
we propose a single model that encodes information from multiple domains. Our model is designed such that test examples from new unknown domains can trigger the most relevant parameters in the model. This way we allow our model to share information between domains and use the most relevant information at test time.  
Our model is inspired by recent research on prompting mechanisms for autoregressive language models. 
We start (\S\ref{sec:DRF-gen}) by describing the general architecture of our model, and continue (\S\ref{sec:DRF}) with the domain related features that form our prompts.

\subsection{The Model}
\label{sec:DRF-gen}
\begin{figure}
\includegraphics[scale=0.2]{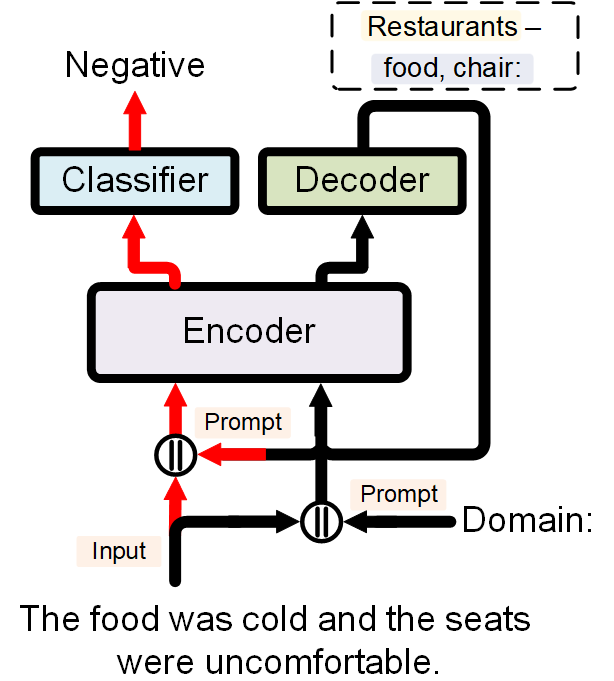}
\centering
\caption{\emph{PADA} during test time inference. An autoregressive model with a generative head trained for DRF generation and a discriminative head for sentiment classification. \emph{PADA} conditions the classification on the generated prompt. Text marked with \colorbox{my_blue}{blue} signifies the DRFs and text marked with \colorbox{my_yellow}{yellow} signifies the domain name. Black arrows ($\rightarrow$) mark the first inference step and \textcolor{red}{red arrows} ($\textcolor{red}{\rightarrow}$) mark the second inference step. 
}
\centering
\label{fig:PADA}
\end{figure}

We present our example-based autoregressive Prompt learning algorithm for on-the-fly Any-Domain Adaptation (\emph{PADA}, Figure \ref{fig:PADA}). \emph{PADA} employs a pre-trained T5 language model and learns to generate example-specific Domain Related Features (DRFs) in order to facilitate accurate task predictions. This is implemented through a two-step multi-task mechanism, where first a DRF set is generated to form a prompt, and then the task label is predicted.

Formally, assume an input example $(x_i,y_i) \sim S_i$, such that $x_i$ is the input text, $y_i$ is the task label and $S_i$ is the domain of this example. For the input $x_i$, \emph{PADA} is trained to first generate $N_i$, the domain name, followed by $R_i$, the DRF signature of $x_i$, and given this prompt to predict the label $y_i$. At test time, when the model encounters an example from an unknown domain, it generates a prompt that may consist of one or more domain names as well as features from the DRF sets of one or more source domains, and based on this prompt it predicts the task label.

\paragraph{Test-time Inference} 
Consider the example in Figure \ref{fig:any-domain}, which describes a sentiment classification model, trained on the \textit{restaurants}, \textit{home-furniture}, \textit{electronic-devices}, and \textit{movies} source domains. The model observes a test example from the \textit{airlines} domain, a previously unseen domain whose name is not known to the model. The model first generates the name of the domain which is most appropriate for this example, \textit{restaurants} in this case. Then, it continues to generate the words ``food'' and ``chair'', features related to the \textit{restaurants} and \textit{home-furniture} domains, respectively. Finally, given this prompt, the model predicts the example's (negative) sentiment.

\paragraph{Training}
In order to separate the prompt generation task from the discriminative classification task, we train our model within a multi-task framework.
\emph{PADA} is trained to perform two tasks, one for generating a prompt, consisting of features from the DRF set of the example's domain, and another for predicting the example's label. For the first, generative task, the model receives examples with the special prompt `\textit{Domain:}', which primes the model to generate $N_i$ and $R_i$ (see examples for prompts generated by \emph{PADA} in Table \ref{tab:drf-gen-examples}). Note that $R_i$ is a set of features derived from the DRF set of $S_i$, and training examples are automatically annotated with their $R_i$, as  described in \S\ref{sec:DRF}. For the second, discriminative task, the model receives a prompt, consisting of $N_i$ and $R_i$, and its task is to predict $y_i$. 

Following the multi-task training protocol of T5, we mix examples from each task. To this end, we define a task proportion mixture parameter $\alpha$. Each example from the training set forms an example for the generative task with probability $\alpha$, and an example for the discriminative task with probability $1 - \alpha$. The greater the value of $\alpha$, the more the model will train for the generative task.

\begin{table}
\small
\centering
\begin{tabular}{p{7cm}}
\toprule
\textbf{Input.} \hspace{0.05cm}
\textit{A sad day for France, for journalism, for free speech, and those whose beliefs the attackers pretend to represent.} \\
\textbf{Prompt.} 
\textit{Ottawa Shooting - french, attack, shootings} \\

\toprule
\textbf{Input.} 
\textit{Picture of explosion in Dammartin-en-Goele. all happened suddenly.} \\
\textbf{Prompt.} 
\textit{Germanwings Crash - explosion, goele, german} \\

\toprule
\textbf{Input.} \hspace{0.05cm}
\textit{At least 5 hostages in kosher supermarket in eastern Paris, according to reports.}\\
\textbf{Prompt.} 
\textit{Paris Siege - hostages, reports, taker} \\

\bottomrule
\end{tabular}
\caption{Examples for DRF-based prompts generated by \emph{PADA}, from the Charlie-Hebdo (C) target domain, which is unknown to \emph{PADA} during training (source domains are FR, GW, OS, and S. See Table \ref{tab:data-stats}). \emph{PADA} generates prompts which are semantically related to the input example by combining DRFs from source domains along with non-DRF yet relevant words. Moreover, it can also generate new domain names (\textit{Paris Siege}).}
\label{tab:drf-gen-examples}
\end{table}

At the heart of our method is the clever selection of the DRF set of each domain, and the prompt annotation process for the training examples. We next discuss these features and their selection process.

\subsection{Domain Related Features}
\label{sec:DRF}
For each domain we define the DRF set such that these features provide a semantic signature for the domain. Importantly, if two domains have shared semantics, for example the \textit{restaurants} and the \textit{cooking} domains, we expect their DRFs to semantically overlap. Since the prompt of each training example consists of a subset of features from the DRF set of its domain, we should also decide on a prompt generation rule that can annotate these training examples with their relevant features.

In order to reflect the semantics of the domain, DRFs should occur frequently in this domain. Moreover, they should be substantially more common in that specific domain relative to all other domains. Despite their prominence in a specific domain, DRFs can also relate to other domains. 
For instance, consider the top example presented in Table~\ref{tab:drf-gen-examples}. The word ``attack'' is highly associated with the ``Ottawa Shooting'' domain and is indeed one of its DRFs. However, this word is also associated with ``Sydney Siege'', which is another domain in the Rumour Detection dataset \citep{zubiaga2016analysing}. Moreover, since both domains are related to similar events, it is not surprising that the DRF set of the former contains the feature \textit{suspect} and the DRF set of the latter contains the feature \textit{taker} (see Table \ref{tab:drf_examples}). The similarity of these features facilitates parameter sharing in our model.

\paragraph{Automatically Extracting DRFs}
There can be several ways of implementing a DRF extraction method that are in line with the above DRF definition.
We experimented with several different extraction criteria (Correlation, class-based TF-IDF,\footnote{\url{https://github.com/MaartenGr/cTFIDF}} and Mutual Information), and observed high similarity ($82\%$ overlap) between their resulting DRF sets. However, we observed a qualitative advantage for Mutual Information (MI), which successfully extracted DRFs that hold domain-specific semantic meaning.

We present the following MI-based method:
Let examples (texts) from the $j$th source domain ($\mathcal{S}_j$) be labeled with $1$, and examples from all other domains ($\mathcal{S} \backslash \mathcal{S}_j$) be labeled with $0$. We first calculate the \textit{mutual-information} (MI) between all tokens and this binary variable, and choose the $l$ tokens with the highest MI score. Note, that the MI criterion might promote tokens which are highly associated with ($\mathcal{S} \backslash \mathcal{S}_j$) rather than with $\mathcal{S}_j$. Thus, we filter the $l$ tokens according to the following condition:
$$ \frac{C_{\mathcal{S}\backslash \mathcal{S}_j}(n)}{ C_{\mathcal{S}_j}(n)} \leq \rho, \;\;  C_{\mathcal{S}_j}(n) > 0 $$
where $C_{\mathcal{S}_j}(n)$ is the count of the n-gram $n$ in $\mathcal{S}_j$,  $C_{\mathcal{S}\backslash \mathcal{S}_j}(n)$ is the count of this n-gram in all source domains except for $\mathcal{S}_j$, and $\rho$ is an n-gram frequency ratio hyper-parameter. 

Intuitively, the smaller $\rho$ is, the more certain we are that the n-gram is especially associated with $\mathcal{S}_j$, compared to other domains. Since the number of examples in $\mathcal{S}_j$ is much smaller than the number of examples in $\mathcal{S} \backslash \mathcal{S}_j$, we choose $\rho \geq 1$ but do not allow it to be too large. As a result, this criterion allows for features which are associated with $\mathcal{S}_j$ but also related to other source domains, to be part of the DRF set of $\mathcal{S}_j$. This is demonstrated in Table~\ref{tab:drfs_by_rhos_examples}, where we present examples of DRFs extracted for the \textit{Ferguson} domain of the rumour detection task, by using different values of $\rho$. Using $\rho=0$, domain-specific DRFs such as ''mikebrown'' are extracted for the domain's DRF set. By increasing the value of $\rho$ to $1$, we add DRFs which are highly associated with the domain, but are also prevalent in other domains (e.g., ''killing'' is also related to the \textit{Ottawa-shooting} domain). However, when increasing the value of $\rho$ to $10$, we extract DRFs which are less associated with the domain (''know''). This is further exacerbated when increasing $\rho$ to higher values.



\begin{table}
\centering
\begin{adjustbox}{width=0.4\textwidth}
\begin{tabular}{ c | c | c | c }
$\rho=0$ & $\rho=1$ & $\rho=10$ & $\rho=100$ \\
\hline
\textit{ferguson} & \textit{police} & \textit{know} & \textit{breaking} \\
\textit{mikebrown}  & \textit{officer} & \textit{report} &  \\
\textit{robbery}  & \textit{killing} & \textit{just} &  \\
\end{tabular}
\end{adjustbox}
\caption{A sample of DRFs extracted for the Ferguson domain (rumour detection) with different $\rho$ values. Each column represents DRFs which are filtered in DRF sets of lower $\rho$ value. DRFs of lower $\rho$ values are more domain specific.}
\label{tab:drfs_by_rhos_examples}
\end{table}

\paragraph{Annotating DRF-based Prompts for Training}
We denote the DRF set of the $j$th domain with $R_j$. Given a training example $i$ from domain $j$, we select the $m$ features from $R_j$ which are most associated with this example to form its prompt. To do that, we compute the Euclidean distance between the T5 embeddings of the DRF features and the T5 embeddings of each of the example's tokens. We then rank this list of pairs by their scores and select the top $m$ features.\footnote{In this computation we consider the non-contextual embeddings learned by T5 during its pre-training. In our experiments we consider only unigrams (words) as DRFs.}
In Table~\ref{tab:drf_examples} we provide a sample of DRFs from the DRF sets associated with each domain in the rumor detection task (\S~\ref{sec:experiments}), alongside their frequency statistics for being annotated in a training example's prompt.

To conclude, our methods for domain-specific DRF set extraction and for prompt annotation of training examples, demonstrate three attractive properties. First, every example has its own unique prompt. Second, our prompts map each training example to the semantic space of its domain. Lastly, the domain-specific DRF sets may overlap in their semantics, either by including the same tokens or by including tokens with similar meanings. This way they provide a more nuanced domain signature compared to the domain name alone. This is later used during the inference phase when the model can generate an example-specific prompt that consists of features from the DRF sets of the various source domains.

%% file: 5.experimental-setup.tex
\section{Experimental Setup}
\label{sec:experiments}
\subsection{Task and Datasets}
\label{sec:datasets}

We experiment with three multi-source DA tasks, where a model is trained on several domains and applied to a new one. 
We consider two text classification tasks, Rumour Detection and Multi-Genre Natural Language Inference (MNLI),  and one sequence tagging task -- Aspect Prediction. The details of the training, development and test sets of each domain are provided in Table~\ref{tab:data-stats}. Our experiments are performed in a leave-one-out fashion: We train the model on all domains but one, and keep the held-out domain for testing. Particularly, training is done on the training data of the source domains and development on their development data, while the test data is taken from the target domain, which is unknown at training time. We repeat the experiments in each task such that each domain is used as a target domain.

\paragraph{Rumour Detection}
The PHEME dataset of rumourous tweets \citep{zubiaga2016analysing,DBLP:conf/socinfo/ZubiagaLP17} contains $5,802$ tweets, which followed $5$ different real-world events, and are labelled as rumourous or non-rumourous.\footnote{\url{https://figshare.com/articles/dataset/PHEME_dataset_of_rumours_and_non-rumours/4010619}} We treat each event as a separate domain: Charlie-Hebdo (C), Ferguson (FR), Germanwings-crash (GW), Ottawa-shooting (OS), and Sydney-siege (S).

We follow the data processing procedure of \citet{wright2020transformer} and split each domain (event) corpus by a $4$:$1$ ratio, establishing training and development sets. Since the corpora are relatively small, we want to avoid further shrinking the size of the test set. Hence, we include all examples available from the target domain to form the test set.\footnote{This does not harm the integrity of our experiments, since the training and development sets are sampled from the source domains while the test set is sampled only from the target domain.}

\paragraph{MNLI}
 This corpus \citep{N18-1101} is an extension of the SNLI dataset \citep{bowman2015large}.\footnote{\url{https://cims.nyu.edu/~sbowman/multinli/}} Each example consists of a pair of sentences, a premise and a hypothesis. The relationship between the two may be entailment, contradiction, or neutral. The corpus includes data from $10$ domains: $5$ are matched, with training, development and test sets, and $5$ are mismatched, without a training set. We experiment only with the five matched domains: Fiction (F), Government (G), Slate (SL), Telephone (TL) and Travel (TR).

Since the test sets of the MNLI dataset are not publicly available, we use the original development sets as our test sets for each target domain, while source domains use these sets for development. 
We explore a lightly supervised scenario, which emphasizes the need for a DA algorithm. Thus, we randomly downsample each of the training sets by a factor of $30$, resulting in $2,000-3,000$ examples per set.

\paragraph{Aspect Prediction}
The Aspect Prediction dataset is based on  aspect-based sentiment analysis (ABSA) corpora from four domains: Device (D), Laptops (L), Restaurant (R), and Service (SE). The D data consists of reviews from \citet{DBLP:conf/acl/ToprakJG10}, the SE data includes web service reviews \citep{DBLP:conf/kdd/HuL04}, and the L and R domains consist of reviews from the SemEval-2014 ABSA challenge \citep{DBLP:conf/semeval/PontikiGPPAM14}.

We follow the training and test splits defined by \citet{DBLP:conf/emnlp/GongYX20} for the D and SE domains, while the splits for the L and R domains are taken from the SemEval-2014 ABSA challenge. 
To establish our development set, we randomly sample $10\%$ out of the training data. 

\begin{table}
\centering
\begin{adjustbox}{width=0.48\textwidth}
\begin{tabular}{c | c | c | c}
\textbf{C}  & \textbf{GW} & \textbf{OS} & \textbf{S} \\
\hline
hebdo	(88\%)  & lufthansa	(86\%) & ottawa	(83\%) & australians	(75\%) \\
ahmed	(48\%)  & germanwings	(33\%) & cdnpoli	(36\%) & monis	(69\%)  \\
terrorists	(22\%)   & crash	(25\%) & shooting	(30\%) &  isis	(21\%)\\
attack	(19\%)   & plane	(24\%) & soldier	(12\%) & cafe	(18\%) \\
victims	(4\%)  & barcelona	(23\%) & suspect	(5\%) &  taker	(16\%) \\
\end{tabular}
\end{adjustbox}
\caption{A sample of DRFs from four rumour detection domains along with their frequency for being annotated in a training example's prompt.}
\label{tab:drf_examples}
\end{table}

\begin{table}[t]
\centering
\begin{adjustbox}{width=0.47\textwidth}
\begin{tabular}{l | c | c | c }
\hline
   \multicolumn{4}{c}{\textbf{Rumour Detection}} \\
   \hline
   \textbf{Domain} & \textbf{Training (src)}  & \textbf{Dev (src)} & \textbf{Test (trg)} \\
   
   \hline
   \textbf{Charlie-Hebdo (C)} & $1,663$ & $416$ & $2,079$ \\
   \textbf{Ferguson (FR)} & $914$ & $229$ & $1,143$ \\
   \textbf{Germanwings-crash (GW)} & $375$ & $94$ & $469$ \\
   \textbf{Ottawa-shooting (OS)} & $712$ & $178$ & $890$ \\
   \textbf{Sydney-siege (S)} & $976$ & $245$ & $1,221$ \\
  \hline
   \multicolumn{4}{c}{\textbf{MNLI} }\\
   \hline
   \textbf{Domain} & \textbf{Training (src)}  & \textbf{Dev (src)} & \textbf{Test (trg)} \\
   
   \hline
   \textbf{Fiction (F)} & $2,547$ & $1,972$ & $1,972$ \\
   \textbf{Government (G)} & $2,541$ & $1,944$ & $1,944$ \\
   \textbf{Slate (SL)} & $2,605$ & $1,954$ & $1,954$ \\
   \textbf{Telephone(TL)} & $2,754$ & $1,965$ & $1,965$ \\
   \textbf{Travel (TR)} & $2,541$ & $1,975$ & $1,975$ \\
   \hline
   \multicolumn{4}{c}{\textbf{Aspect} }\\
   \hline
   \textbf{Domain} & \textbf{Training (src)}  & \textbf{Dev (src)} & \textbf{Test (trg)} \\
   
   \hline
   \textbf{Device (D)} & $2,302$ & $255$ & $1,279$ \\
   \textbf{Laptops (L)} & $2,726$ & $303$ & $800$ \\
   \textbf{Restaurants (R)} & $3,487$ & $388$ & $800$ \\
   \textbf{Service(SE)} & $1,343$ & $149$ & $747$ \\

\end{tabular}
\end{adjustbox}
\caption{The number of examples in each domain of our three tasks. We denote the examples used when a domain is included as a source domain (src), and when it is the target domain (trg).}
\label{tab:data-stats}
\end{table}

\subsection{Evaluated Models}
\label{sec:models-and-baselines}

Our main model is \textbf{\emph{PADA}}: The  multi-task model that first generates the domain name and domain related features to form a prompt, and then uses this prompt to predict the task label (\S\ref{sec:DRF-gen}, Figure~\ref{fig:PADA}).
We compare it to two types of models:  (a) T5-based baselines corresponding to ideas presented in multi-source DA work, as well as other recent state-of-the-art models (\S\ref{sec:related}); and (b) Ablation models that use specific parts of  \emph{PADA}, to highlight the importance of its components. 




\subsubsection{Baseline Models}
\label{sec:baseline models}


\paragraph{\emph{Transformer-based Mixture of Experts (Tr-MoE)}}
For each source domain, a separate transformer-based DistilBERT expert model \citep{DBLP:journals/corr/abs-1910-01108} is trained on the domain's training set, and an additional model is trained on the union of training sets from all source domains. At test time, the average of the class probabilities of these models is calculated and the highest probability class is selected. This model is named \emph{MoE-avg} by \citet{wright2020transformer} and has demonstrated to achieve state-of-the-art performance for Rumour Detection.

\paragraph{\emph{T5-MoE}} 
A T5-based MoE ensemble model. For each source domain, a separate pre-trained T5 model is fine-tuned on the domain's training set (i.e. a domain expert model). During inference, the final predictions of the model are decided using the same averaging procedure as in \emph{Tr-MoE}.

\paragraph{\emph{T5-No-Domain-Adaptation (T5-NoDA)}} 
A pre-trained T5 model, which feeds the same task classifier used in \emph{PADA} (see below) to predict the task label. In each DA setting, the model is trained on the training data from all source domains.

We also experiment with an in-domain version of this model, \textbf{\emph{T5-UpperBound (T5-UB)}}, which is tested on the development data of each domain. We treat \emph{T5-UB} performance as an upper bound for the average target performance across all DA settings, for any T5-based model in our setup.

\paragraph{\emph{T5-Domain-Adversarial-Network (T5-DAN)}} 
A model that integrates \emph{T5-NoDA} with an adversarial domain classifier to learn domain invariant representations.\footnote{We also experimented with \emph{BERT-NoDA} and \emph{BERT-DAN} models. We do not report their results since they were consistently outperformed by \emph{T5-NoDA} and \emph{T5-DAN}.} 

\paragraph{\emph{T5-Invariant-Risk-Minimization (T5-IRM)}} 
A T5-based model which penalizes feature distributions that have different optimal linear classifiers for each domain. The model is trained on the training data from all source domains.

IRM \citep{DBLP:journals/corr/abs-1907-02893} and DAN \citep{DBLP:journals/jmlr/GaninUAGLLML16} are established algorithms in the domain robustness literature, for generalization to unseen distributions \citep{DBLP:journals/corr/abs-2012-07421}.  


\subsubsection{Ablation Models}
\label{sec:ablation models}

\paragraph{\emph{Prompt-DN}}
A simplified version of our \emph{PADA} model, which assigns only a \textit{domain name} as a prompt to the input text. Since the domain name is unknown at test time, we create multiple variants of each test example, each with one of the training domain names as a prompt. For the final predictions of the model we follow the same averaging procedure as in \emph{Tr-MoE} and \emph{T5-MoE}. 

\paragraph{\emph{Prompt-RDW} and \emph{Prompt-REW}}
Two simplified versions of \emph{PADA} which form prompts from \emph{Random-Domain-Words} and \emph{Random-Example-Words}, respectively. For \emph{Prompt-RDW}, we sample $m=5$ domain words (according to their distribution in the joint vocabulary of all source domains) for each example. For \emph{Prompt-REW}, we randomly select $m=5$ words from the example's text.
At both training and test times, we follow the same prompt formation procedures.

\begin{figure}[ht]
\begin{subfigure}{0.23\textwidth}
\includegraphics[scale=0.18]{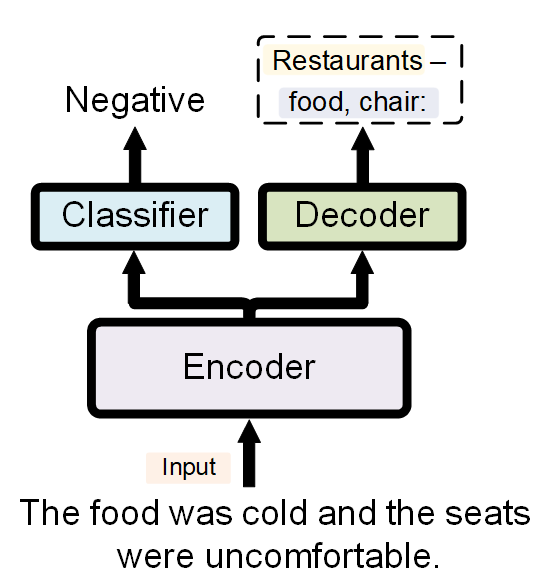}
\centering
\caption{\emph{PADA-NP}.}

\centering
\label{fig:PADA-NP}
\end{subfigure}
\begin{subfigure}{0.23\textwidth}
\includegraphics[scale=0.18]{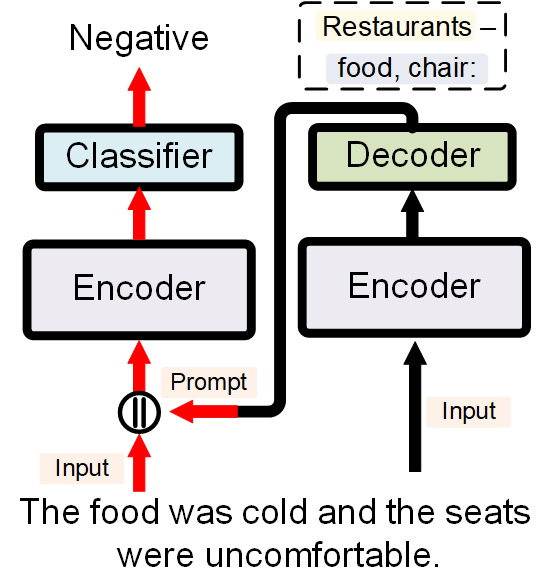}
\centering
\caption{\emph{PADA-NM}.}
\centering
\label{fig:PADA-NM}
\end{subfigure}
\label{fig:pada-ablations}
\caption{\emph{PADA} ablation models: (a) \emph{PADA-NP}, which follows a multi-task training protocol, but does not condition its prediction on the generated prompt; (b) \emph{PADA-NM}, which separately trains a prompt generation model ($\textcolor{black}{\rightarrow}$) and a prompted task prediction model ($\textcolor{red}{\rightarrow}$).}
\end{figure}

\paragraph{\emph{PADA-NP (No Prompt)}}
A multi-task model similar to \emph{PADA}, except that it simultaneously generates the example-specific domain name and DRF-based prompt, and predicts the task label (Figure \ref{fig:PADA-NP}). 
Since this model does not condition the task prediction on the generated prompt, it sheds light on the effect of the autoregressive nature of \emph{PADA}.

\paragraph{\emph{PADA-NM (No Multi-task)}}
A pipeline of two independent models which emulates \emph{PADA}. Given an input example, the first model generates a unique prompt for it. Then, the second model predicts the task label given the input and its generated prompt (Figure \ref{fig:PADA-NM}). 
Since the prediction and prompt generation tasks are not performed jointly, nor are the model parameters shared between the tasks, this pipeline sheds light on the effect of the multi-task nature of \emph{PADA}.

\subsection{Implementation Details}
For all implemented models we use the \emph{HuggingFace Transformers} library \citep{wolf-etal-2020-transformers}.\footnote{\url{https://github.com/huggingface/transformers}}

The T5-based text classification models do not follow the same procedure originally described in \citet{raffel2020exploring}. Instead, we add a simple \emph{1D-CNN} classifier on top of the T5 encoder to predict the task label (Figure \ref{fig:PADA}). The number of filters in this classifier is $32$ with a filter size of $9$.\footnote{We experimented with the original T5 classification method as well, but \emph{PADA} consistently outperformed it.} The generative component of the T5-based models is identical to that of the original T5.
Our T5-based models for Aspect Prediction cast sequence tagging as a sequence-to-sequence task, employing the  text-to-text approach of \citet{raffel2020exploring} to generate a `B' (begin), `I' (in) or `O' (out) token for each input token. Other than this change, these models are identical to the T5-based models for text classification.

We train all text classification models for $5$ epochs and all sequence tagging models for $60$ epochs, with an early stopping criterion according to performance on the development data. We use the cross-entropy loss function for all models, optimizing their parameters with the \emph{ADAM} optimizer \citep{DBLP:journals/corr/KingmaB14}. We employ a batch size of $32$ for text classification and $24$ for sequence tagging, warmup ratio of $0.1$, and a learning rate of $5\cdot10^{-5}$. The maximum input and output lengths of all T5-based models is set to $128$ tokens. We pad shorter sequences and truncate longer ones to the maximum input length.

For \emph{PADA}, we tune the $\alpha$ (example proportion-mixture, see \S\ref{sec:DRF-gen}) parameter considering the value range of $\{0.1,0.25,0.5,0.75,0.9\}$. The chosen values are: $\alpha_{rumour}=0.75$, $\alpha_{mnli}=0.1$ and $\alpha_{absa}=0.1$. For each training example, we select the top $m=5$ DRFs most associated with it for its prompt. For the generative component of the T5-based models, we perform inference with the Diverse Beam Search algorithm \citep{DBLP:journals/corr/VijayakumarCSSL16}, considering the following hyper-parameters: We generate $5$ candidates, using a beam size of $10$, with $5$ beam groups, and a diversity penalty value of $1.5$. The $l$ and $\rho$ parameters of the DRF extraction procedure (\S\ref{sec:DRF}) were tuned to $1000$ and $1.5$, respectively, for all domains.

%% file: 6.results.tex
\begin{table*}
\centering
\begin{adjustbox}{width=\textwidth}
\begin{tabular}{| l || c | c | c | c | c || c || c | c | c | c | c || c |}
\toprule
   & \multicolumn{6}{c||}{\textbf{Rumour Detection}} & \multicolumn{6}{c|}{\textbf{MNLI}} \\
  \midrule
   & \textbf{All $\rightarrow$ C} & \textbf{All $\rightarrow$ FR} & \textbf{All $\rightarrow$ GW} & \textbf{All $\rightarrow$ OS} & \textbf{All $\rightarrow$ S} & \textbf{AVG} & \textbf{All $\rightarrow$ F} & \textbf{All $\rightarrow$ G} & \textbf{All $\rightarrow$ SL} & \textbf{All $\rightarrow$ TE} & \textbf{All $\rightarrow$ TR} & \textbf{AVG} \\
   \midrule
    
     
     \emph{Tr-MoE} & $ 68.0 $ & $ 46.1 $ & $ 74.8 $ & $ 58.2 $ & $ 64.9 $ & $ 62.4 $ & $ 64.3 $ & $ 73.9 $ & $ 65.3 $ & $ 62.4 $ & $ 69.8 $ & $ 67.1 $ \\
     
     \emph{T5-MoE} & $ 68.1 $ & $ 46.0 $ & $ 73.6 $ & $ 65.3 $ & $ 66.3 $ & $ 63.9 $ & $ 74.0 $ & $ 82.0 $ & $ 73.4 $ & $ 74.6 $ & $ 78.3 $ & $ 76.5 $ \\
     

     \emph{T5-DAN} & $ 64.9 $ & $ 52.4 $ & $ 69.1 $ & $ 72.7 $ & $ 64.4 $ & $ 64.7 $ & $ 74.4 $ & $ 76.3 $ & $ 61.0 $ & $ 72.4 $ & $ 77.7 $ & $ 72.4 $ \\
     
     \emph{T5-IRM} & $ 63.5 $ & $ 39.4 $ & $ 70.1 $ & $ 44.2 $ & $ 65.7 $ & $ 56.6 $ & $ 72.0 $ & $ 81.5 $ & $ 73.2 $ & $ 69.3 $ & $ 78.9 $ & $ 75.0 $ \\
     
     \emph{T5-NoDA} & $ 64.1 $ & $ 46.9 $ & $ \textbf{75.1} $ & $ 72.0 $ & $ 71.0 $ & $ 65.8 $ & $ 76.4 $ & $ 83.5 $ & $ 75.5 $ & $ 74.9 $ & $ 81.3 $ & $ 78.3 $ \\
     
     \midrule
     
     \emph{Prompt-DN} & $ 66.4 $ & $ 53.7 $ & $ 72.4 $ & $ 71.4 $ & $ 70.1 $ & $ 66.8 $ & $ \textbf{77.0} $ & $ \textbf{84.4} $ & $ 75.6 $ & $ 76.3 $ & $ 80.5 $ & $ 78.8 $ \\
     
     \emph{Prompt-RDW} & $ 64.1 $ & $ 53.1 $ & $ 71.8 $ & $ 66.0 $ & $ 70.0 $ & $ 65.0 $ & $ 76.0 $ & $ 84.2 $ & $ 76.6 $ & $ 77.0 $ & $ 79.9 $ & $ 78.7 $ \\
     
     \emph{Prompt-REW} & $ 64.2 $ & $ 54.3 $ & $ 71.6 $ & $ 70.0 $ & $ 69.1 $ & $ 65.8 $ & $ 75.7 $ & $ 81.4 $ & $ 76.7 $ & $ 78.8 $ & $ 81.2 $ & $ 78.7 $ \\
     
     \midrule
     
     \emph{PADA-NP} & $ 65.8 $ & $ \textbf{54.8} $ & $ 71.6 $ & $ 72.2 $ & $ 74.0 $ & $ 67.7 $ & $ 76.2 $ & $ 83.6 $ & $ 75.4 $ & $ 77.2 $ & $ 81.4 $ & $ 78.8 $ \\
     
     \emph{PADA-NM} & $ 63.6 $ & $ 54.1 $ & $ 74.3 $ & $ 70.1 $ & $ 70.3 $ & $ 66.5 $ & $ 76.0 $ & $ 83.7 $ & $ 76.5 $ & $ 78.0 $ & $ 81.0 $ & $ 79.0 $ \\
     
     \midrule
     
     \emph{PADA} & $ \textbf{68.6} $ & $ 54.4 $ & $ 73.0 $ & $ \textbf{75.2} $ & $ \textbf{75.1} $ & $\textbf{69.3}$ & $ 76.4 $ & $ 83.4 $ & $ \textbf{76.9} $ & $ \textbf{78.9} $ & $ \textbf{82.5} $ & $ \textbf{79.6} $ \\ 
    \bottomrule      
\end{tabular}
\end{adjustbox}
\caption{Binary-F1 scores for the Rumour Detection task and macro-F1 scores for the MNLI task.}
\label{tab:rumour and mnli}
\end{table*}

\section{Results}
\label{sec:results}

\begin{table}[ht!]
\centering
\begin{adjustbox}{width=0.48\textwidth}
\begin{tabular}{| l || c | c | c | c || c |}
  \toprule
  \multicolumn{6}{|c|}{\textbf{Aspect Prediction}} \\
  \midrule
  & \textbf{All $\rightarrow$ D} & \textbf{All $\rightarrow$ L} & \textbf{All $\rightarrow$ R} & \textbf{All $\rightarrow$ SE} & \textbf{AVG} \\
  \midrule
    
     \emph{T5-MoE} &  $ 39.5 $  & $ 31.4 $ & $ 31.4 $ & $ 30.9 $ & $ 33.3 $ \\
     

     \emph{T5-DAN} &  $ 28.4 $  & $ 38.0 $ & $ 49.1 $ & $ 33.4 $ & $ 33.2 $ \\
     
     \emph{T5-IRM} &  $ 37.1 $  & $ 44.6 $ & $ 47.4 $ & $ 41.5 $ & $ 42.7 $ \\
     
     \emph{T5-NoDA} &  $ 31.1 $  & $ 45.6 $ & $ 40.2 $ & $ 37.9 $ & $ 38.7 $ \\
     
     \midrule
     
     \emph{Prompt-DN} &  $ 41.1 $ & $ 42.6 $ & $ 29.0 $ & $ 30.8 $ & $ 35.9 $ \\
     
     \emph{Prompt-RDW} &  $ 34.6 $ & $ 46.9 $ & $ \textbf{52.9} $ & $ 41.2 $ & $ 43.9 $ \\
     
     \emph{Prompt-REW} &  $ 38.2 $ & $ 49.5 $ & $ 45.1 $ & $ 39.6 $ & $ 43.1 $ \\
     
     \midrule
     
     \emph{PADA-NP} &  $ 41.7 $ & $ 48.2 $ & $ 50.1 $ & $ 40.1 $ & $ 45.0 $ \\
     
     \emph{PADA-NM} &  $ 40.3 $ & $ 48.8 $ & $ 50.8 $ & $ 40.2 $ & $ 45.0 $ \\
     
     \midrule
     
     \emph{PADA} &  $ \textbf{43.1} $ & $ \textbf{50.9} $ & $ 50.8 $ & $ \textbf{45.3} $ & $ \textbf{47.5} $ \\ 
    \bottomrule      
\end{tabular}
\end{adjustbox}
\caption{Binary-F1 scores for Aspect Prediction.}
\label{tab:absa}
\end{table}

\paragraph{Text Classification}

Table \ref{tab:rumour and mnli} presents our results.  We report the binary-F1 score for Rumour Detection, and the macro-F1 score for MNLI.\footnote{Binary-F1 measures the F1 score of the positive class. It is useful in cases of unbalanced datasets where the positive class is of interest ($34\%$ of the Rumour Detection dataset).}
%
\emph{PADA} outperforms all baseline models (\S~\ref{sec:baseline models}) in 7 of 10 settings and reaches the highest result in another setting (with \textit{T5-NoDA}), exhibiting average performance gains of $3.5\%$ and $1.3\%$ in Rumour Detection and MNLI, respectively, over the best performing baseline model. Interestingly, it is \textit{T5-NoDA}, which does not perform any DA, that outperforms (on average and in most model-to-model comparisons) all other baseline model, including the MoE models.

While the performance gains differ between the tasks, they partly stem from the different performance gaps between source and target domains in each of these tasks. Recall, that we consider the \textit{T5-UB} performance on its development sets for Rumour Detection ($82.8\%$) and MNLI ($80.8\%$), to be the upper bound for the average target performance across all DA settings, for any T5-based model.
When considering the gaps between this upper bound and \textit{T5-NoDA} ($65.8\%$ for Rumour Detection and $78.3\%$ for MNLI), \emph{PADA} reduces the error rate by $21\%$ for Rumour Detection and $52\%$ for MNLI. The improvements gained by \emph{PADA} are in fact substantial in both tasks.

The advantage of \emph{PADA} over MoE goes beyond improved predictions. Particularly, for \emph{PADA} we train a single model while for MoE we train a unique model for each source domain, hence the number of parameters in the MoE framework linearly increases with the number of source domains. For example, in our setups, \textit{Tr-MoE} trains five DistilBERT models (one for each source domain and one for all source domains together), resulting in $5 \cdot 66M = 330M$ parameters. In contrast, the \emph{PADA} models keep the $220M$ parameters of T5, regardless of the number of source domains.

\paragraph{Sequence Tagging} In order to demonstrate the wide applicability of our approach, we go beyond text classification (with 2 (Rumour Detection) or 3 (MNLI) classes) and also consider Aspect Prediction: A sequence tagging task. We are particularly curious to see if the aforementioned patterns replicate in this qualitatively different task.
Our results are presented in Table \ref{tab:absa}, where we report the binary-F1 score (the F1 score of the aspect class). Crucially, the patterns we observe for text classification can also be detected for sequence tagging. Particularly, \emph{PADA} is the best performing model in 4 of 4 settings compared to its baselines. On average, \emph{PADA} outperforms the second-best model, \textit{T5-IRM}, by $3.5\%$ on average. Given the average results of \textit{T5-UB} ($69.4\%$) and \textit{T5-NoDA} ($38.7\%$), the error reduction is $24\%$.

\paragraph{PADA Ablation Models}
As shown in Table~\ref{tab:rumour and mnli}, \emph{PADA} outperforms all of its variants (\S~\ref{sec:ablation models}) in 6 out of 10 text classification settings overall. Furthermore, in the sequence tagging task (Table~\ref{tab:absa}), \emph{PADA} outperforms its simpler variants (Prompt-\{DN, REW\}, PADA-NP) in all 4 setups, and Prompt-RDW, PADA-NM in 3 out of 4 setups.
These results highlight the importance of our design choices: (a) including DRFs in the example-specific prompts, tailoring them to express the relation between the source domains and the test example (\emph{PADA} vs \textit{Prompt-\{DN, RDW, REW\}}); (b) utilizing an autoregressive component, where the generated DRF prompts are used by the task classification component (\emph{PADA} vs \textit{PADA-NP}); and (3) leveraging a multi-task training objective (\emph{PADA} vs \textit{PADA-NM}).
A noticeable difference in the aspect prediction results from text classification results is the weakness of \textit{Prompt-DN}, which is outperformed by all baseline models (\S~\ref{sec:baseline models}) in 2 setups, and by 2 of these models in a third setup, as well as on average across all setups. This is yet another indication of the importance of the DRFs in the prompt generated by \emph{PADA}.

%% file: 7.ablation.tex
\section{Ablation Analysis}
\label{sec:ablation}

In this section, we analyze several unique aspects of \emph{PADA}. We first evaluate the prompts generated by \emph{PADA}, to gain further insight into its generative capabilities.
We then analyze the impact of the number of source domains on \emph{PADA}'s performance.
Finally, we examine performance drops due to domain shifts, in order to evaluate \emph{PADA}'s adaptation stability across domains.
For the sake of clarity and concision, analyses will henceforth focus on the rumour detection task.

\paragraph{Generated Prompts Analysis}

\begin{table}
\centering
\begin{adjustbox}{width=0.48\textwidth}
\begin{tabular}{| l | c | c | c | c |}
  \hline
&\textbf{\textit{BERTScore}} & \textbf{\textit{ROUGE-1}}  & \textbf{\textit{ROUGE-2}} & \textbf{\textit{ROUGE-L}} \\
\hline
 \textbf{Dev F1} & $0.94$ &  $ 0.64 $  & $ 0.30 $ & $ 0.61 $  \\
  \hline
  \end{tabular}
\end{adjustbox}
\caption{Average F1 scores for our automatic evaluation metrics, calculated for generated prompts compared to annotated prompts over all development sets in the rumour detection task.}
\label{tab:generated_prompts_metrics}
\end{table}

We first present an intrinsic evaluation of \emph{PADA}'s prompt generation task (see \S\ref{sec:DRF-gen}) by examining model-generated prompts for examples from the development set, compared to their annotated prompts.\footnote{\emph{PADA} is not restricted to specific structures or vocabulary when generating prompts, hence our annotated prompts only serve as pseudo gold labels for training purposes.} We choose automatic metrics widely used for evaluating NLG tasks, focusing on n-gram overlap by calculating ROUGE \cite{Lin2004ROUGEAP} scores as well as measuring semantic similarity with BERTScore \cite{DBLP:conf/iclr/ZhangKWWA20}.
In Table \ref{tab:generated_prompts_metrics} we present average F1 scores for these metrics, calculated over all DA settings in the rumour detection task. 
The high average BERTScore ($0.94$) indicates that the generated prompts share high semantic similarity with their annotated prompts. Yet, the average ROUGE-1 ($0.64$) and ROUGE-2 ($0.3$) scores, indicate that the generated prompts vary on their unigram and bigram levels (respectively), compared with their annotated prompts. This evidence suggests that \emph{PADA} learns to leverage the semantic overlaps between DRFs, over memorizing specific n-grams (f.e. an annotated DRF may be \textit{terrorist} while the generated may be \textit{gunman}).

\begin{figure}
\includegraphics[scale=0.4]{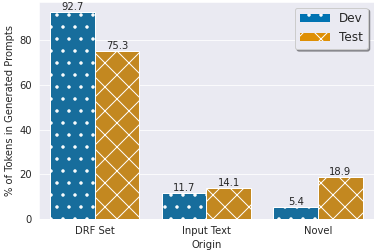}
\centering
\caption{Average token source ratios in generated prompts, calculated over all development and test sets in the rumour detection task.}
\centering
\label{fig:generated_prompts_sources}
\end{figure}

We continue our evaluation by analyzing the origins of words in the \emph{PADA}-generated prompts, specifically, whether they appear in the source domains' DRF sets, the input text, or in neither (Novel).
Figure \ref{fig:generated_prompts_sources} presents the average ratios of different origins for generated prompt tokens, calculated over all DA settings in the rumour detection task. 
As expected, the overwhelming majority of generated tokens come from the source domains DRF sets, for both development ($92.7\%$) and test ($75.3\%$) sets. 
However, when introduced to examples from unknown domains (test sets), we observe a significant increase (compared to the development sets) in novel tokens ($18.9\%$ vs $5.4\%$) and a slight increase in tokens from the example's input text ($14.1\%$ vs $11.7\%$).

\begin{figure}
\includegraphics[scale=0.4]{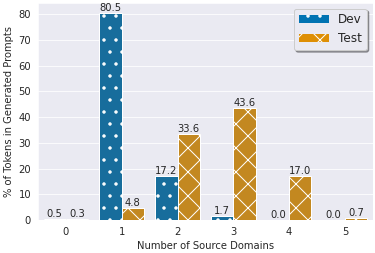}
\centering
\caption{Average ratios of number of domains in generated prompts, calculated over all development and test sets in the rumour detection task.}
\centering
\label{fig:generated_prompts_num_sources}
\end{figure}

Furthermore, Figure \ref{fig:generated_prompts_num_sources} demonstrates that \emph{PADA} is able to exploit information from its multiple source domains. For test examples \emph{PADA} generates prompts containing DRFs from several domains ($95\%$ of prompts contain DRFs from more than $2$ source domains), while for development examples it mostly generates prompts with DRFs only from the correct source domain.
Together with the examples presented in Table \ref{tab:drf-gen-examples}, these observations suggest an encouraging finding - \emph{PADA} is successful in generating prompts which leverage and integrate both the source domains and the semantics of the input example.

\begin{figure}
\includegraphics[scale=0.29]{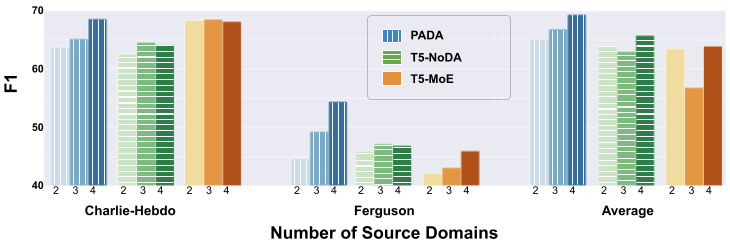}
\centering
\caption{Performance on the Rumour Detection task by the number of source domains the model was trained on. Darker hues represent a larger number of source domains.}
\centering
\label{fig:num-sources}
\end{figure}

\paragraph{Number of Source Domains}
We next turn to study the impact of the number of source domains on \emph{PADA}'s overall performance. Figure~\ref{fig:num-sources} presents F1 scores by the number of source domains for \emph{PADA} and two of its baselines, namely \textit{T5-NoDA} and \textit{T5-MoE}. We provide results on two target domains, as well as an average score across all five target domains from the rumour detection dataset.

As indicated in the figure, \emph{PADA}'s performance improves as the number of source domains increases. These results support our claim that \emph{PADA} is able to integrate knowledge from multiple source domains by learning a meaningful domain-mixture, and it then leverages this knowledge when introduced to an example from a new, unknown, domain. Interestingly, for the baseline models \textit{T5-NoDA} and \textit{T5-MoE}, it seems that including more source domains can sometimes harm their ability to generalize to unknown target domains.
One of our main hypotheses states that a DA model stands to benefit from incorporating combined knowledge from multiple source domains (\S\ref{sec:prompt-based-da}). \emph{PADA} successfully implements this idea, while \textit{T5-MoE} and \textit{T5-NoDA} fall short.



\paragraph{Performance Drops between Source and Target}


When a DA method improves model performance on the target domain, this can result in either increasing or decreasing the performance gap between the source and target domains. If a model performs similarly on its source training domains and on unseen target domains, its source domain performance can also provide an important indication for its future performance in such unseen domains. We hence consider such stability in performance as a desired property in our setup where future target domains are unknown (see discussion in \citet{DBLP:conf/acl/ZiserR19}).

\begin{figure}
\includegraphics[scale=0.35]{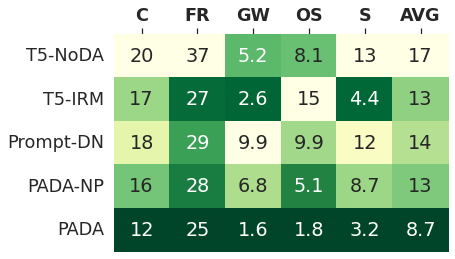}
\centering
\caption{A heatmap presenting performance drops between source domains and target domains (columns), for the rumour detection task. Darker colors represent smaller performance drops.}
\centering
\label{fig:performance-shifts-heatmaps}
\end{figure}

Figure \ref{fig:performance-shifts-heatmaps} presents a heatmap, depicting the performance drop for each model between the source domains and the target domains in rumour detection. 
We measure each model's in-domain performance by calculating an F1 score across all development examples from its source domains, as well as out-of-domain performance on the target domain test set, as described in \S\ref{sec:results}. We then calculate the difference between the source and the target performance measures, and report results for the best performing models in our experiments (\S\ref{sec:results}).
The general trend is clear: \emph{PADA} not only performs better on the target domain, but it also substantially reduces the source-target performance gap. While \textit{T5-NoDA}, which is not a DA model, triggers the largest average absolute performance drop, $17\%$, the average of \emph{PADA}'s absolute performance drop is $8.7\%$.

%% file: 8.discussion.tex
\section{Discussion}


We addressed the problem of multi-source domain adaptation when the target domain is not known at training time. Effective models for this setup can be applied to any target domain with no data requirements about the target domains and without an increase in the number of model parameters as a function of the number of source or target domains. \emph{PADA}, our algorithm, extends the prompting mechanism of the T5 autoregressive language model to generate a unique textual prompt per example. Each generated prompt maps its test example into a semantic space spanned by the source domains.


Our experimental results with three tasks and fourteen multi-source adaptation settings demonstrate the effectiveness of our approach compared to strong alternatives, as well as the importance of the model components and of our design choices. Moreover, as opposed to the MoE paradigm, where a model is trained separately for each source domain, \emph{PADA} provides a single unified model. Intuitively, this approach also seems more cognitively plausible -- a single model attempts to adapt itself to examples from new incoming domains, rather than employing an independent model per domain.

The prompt generation mechanism of \emph{PADA} is naturally limited by the set of source domains it is trained on. This might yield sub-optimal DRFs in prompts generated for examples stemming from target domains which are semantically unrelated to any of the source domains. To alleviate this issue, we allow \emph{PADA} to generate non-DRF words. Still, our prompt generation training process does not directly optimize for the downstream prediction task's objective, which might also contribute to sub-optimally generated prompts.
In future work, we hope to improve these aspects of our approach and explore natural extensions that accommodate multiple tasks and domains in a single model.





%% file: main.bbl
\begin{thebibliography}{73}
\expandafter\ifx\csname natexlab\endcsname\relax\def\natexlab#1{#1}\fi

\bibitem[{Arjovsky et~al.(2019)Arjovsky, Bottou, Gulrajani, and
  Lopez{-}Paz}]{DBLP:journals/corr/abs-1907-02893}
Mart{\'{\i}}n Arjovsky, L{\'{e}}on Bottou, Ishaan Gulrajani, and David
  Lopez{-}Paz. 2019.
\newblock \href {http://arxiv.org/abs/1907.02893} {Invariant risk
  minimization}.
\newblock \emph{CoRR}, abs/1907.02893.

\bibitem[{Ben-David et~al.(2020)Ben-David, Rabinovitz, and
  Reichart}]{ben2020perl}
Eyal Ben-David, Carmel Rabinovitz, and Roi Reichart. 2020.
\newblock \href {https://transacl.org/ojs/index.php/tacl/article/view/2255}
  {Perl: Pivot-based domain adaptation for pre-trained deep contextualized
  embedding models}.
\newblock \emph{Transactions of the Association for Computational Linguistics},
  8:504--521.

\bibitem[{Blitzer et~al.(2007)Blitzer, Dredze, and
  Pereira}]{DBLP:conf/acl/BlitzerDP07}
John Blitzer, Mark Dredze, and Fernando Pereira. 2007.
\newblock \href {https://www.aclweb.org/anthology/P07-1056/} {Biographies,
  bollywood, boom-boxes and blenders: Domain adaptation for sentiment
  classification}.
\newblock In \emph{{ACL} 2007, Proceedings of the 45th Annual Meeting of the
  Association for Computational Linguistics, June 23-30, 2007, Prague, Czech
  Republic}. The Association for Computational Linguistics.

\bibitem[{Blitzer et~al.(2009)Blitzer, Foster, and Kakade}]{blitzer2009zero}
John Blitzer, Dean~P. Foster, and Sham~Machandranath Kakade. 2009.
\newblock \href
  {https://citeseerx.ist.psu.edu/viewdoc/download?doi=10.1.1.147.351&rep=rep1&type=pdf}
  {Zero-shot domain adaptation: A multi-view approach}.

\bibitem[{Blitzer et~al.(2006)Blitzer, McDonald, and
  Pereira}]{DBLP:conf/emnlp/BlitzerMP06}
John Blitzer, Ryan McDonald, and Fernando Pereira. 2006.
\newblock \href {https://www.aclweb.org/anthology/W06-1615/} {Domain adaptation
  with structural correspondence learning}.
\newblock In \emph{{EMNLP} 2006, Proceedings of the 2006 Conference on
  Empirical Methods in Natural Language Processing, 22-23 July 2006, Sydney,
  Australia}, pages 120--128. {ACL}.

\bibitem[{Bowman et~al.(2015)Bowman, Angeli, Potts, and
  Manning}]{bowman2015large}
Samuel Bowman, Gabor Angeli, Christopher Potts, and Christopher Manning. 2015.
\newblock \href {https://doi.org/10.18653/v1/d15-1075} {A large annotated
  corpus for learning natural language inference}.
\newblock In \emph{Proceedings of the 2015 Conference on Empirical Methods in
  Natural Language Processing, {EMNLP} 2015, Lisbon, Portugal, September 17-21,
  2015}, pages 632--642. The Association for Computational Linguistics.

\bibitem[{Brown et~al.(2020)Brown, Mann, Ryder, Subbiah, Kaplan, Dhariwal,
  Neelakantan, Shyam, Sastry, Askell, Agarwal, Herbert{-}Voss, Krueger,
  Henighan, Child, Ramesh, Ziegler, Wu, Winter, Hesse, Chen, Sigler, Litwin,
  Gray, Chess, Clark, Berner, McCandlish, Radford, Sutskever, and
  Amodei}]{DBLP:conf/nips/BrownMRSKDNSSAA20}
Tom~B. Brown, Benjamin Mann, Nick Ryder, Melanie Subbiah, Jared Kaplan,
  Prafulla Dhariwal, Arvind Neelakantan, Pranav Shyam, Girish Sastry, Amanda
  Askell, Sandhini Agarwal, Ariel Herbert{-}Voss, Gretchen Krueger, Tom
  Henighan, Rewon Child, Aditya Ramesh, Daniel~M. Ziegler, Jeffrey Wu, Clemens
  Winter, Christopher Hesse, Mark Chen, Eric Sigler, Mateusz Litwin, Scott
  Gray, Benjamin Chess, Jack Clark, Christopher Berner, Sam McCandlish, Alec
  Radford, Ilya Sutskever, and Dario Amodei. 2020.
\newblock \href
  {https://proceedings.neurips.cc/paper/2020/hash/1457c0d6bfcb4967418bfb8ac142f64a-Abstract.html}
  {Language models are few-shot learners}.
\newblock In \emph{Advances in Neural Information Processing Systems 33: Annual
  Conference on Neural Information Processing Systems 2020, NeurIPS 2020,
  December 6-12, 2020, virtual}.

\bibitem[{Chen et~al.(2012)Chen, Xu, Weinberger, and
  Sha}]{DBLP:conf/icml/ChenXWS12}
Minmin Chen, Zhixiang~Eddie Xu, Kilian~Q. Weinberger, and Fei Sha. 2012.
\newblock \href {http://icml.cc/2012/papers/416.pdf} {Marginalized denoising
  autoencoders for domain adaptation}.
\newblock In \emph{Proceedings of the 29th International Conference on Machine
  Learning, {ICML} 2012, Edinburgh, Scotland, UK, June 26 - July 1, 2012}.
  icml.cc / Omnipress.

\bibitem[{Chen and Cardie(2018)}]{DBLP:conf/naacl/ChenC18}
Xilun Chen and Claire Cardie. 2018.
\newblock \href {https://doi.org/10.18653/v1/n18-1111} {Multinomial adversarial
  networks for multi-domain text classification}.
\newblock In \emph{Proceedings of the 2018 Conference of the North American
  Chapter of the Association for Computational Linguistics: Human Language
  Technologies, {NAACL-HLT} 2018, New Orleans, Louisiana, USA, June 1-6, 2018,
  Volume 1 (Long Papers)}, pages 1226--1240. Association for Computational
  Linguistics.

\bibitem[{Daum{\'e}~III and Marcu(2006)}]{daume2006domain}
Hal Daum{\'e}~III and Daniel Marcu. 2006.
\newblock \href {https://doi.org/10.1613/jair.1872} {Domain adaptation for
  statistical classifiers}.
\newblock \emph{Journal of Artificial Intelligence Research}, 26:101--126.

\bibitem[{Devlin et~al.(2019)Devlin, Chang, Lee, and
  Toutanova}]{DBLP:conf/naacl/DevlinCLT19}
Jacob Devlin, Ming{-}Wei Chang, Kenton Lee, and Kristina Toutanova. 2019.
\newblock \href {https://doi.org/10.18653/v1/n19-1423} {{BERT:} pre-training of
  deep bidirectional transformers for language understanding}.
\newblock In \emph{Proceedings of the 2019 Conference of the North American
  Chapter of the Association for Computational Linguistics: Human Language
  Technologies, {NAACL-HLT} 2019, Minneapolis, MN, USA, June 2-7, 2019, Volume
  1 (Long and Short Papers)}, pages 4171--4186. Association for Computational
  Linguistics.

\bibitem[{Ganin et~al.(2016)Ganin, Ustinova, Ajakan, Germain, Larochelle,
  Laviolette, Marchand, and Lempitsky}]{DBLP:journals/jmlr/GaninUAGLLML16}
Yaroslav Ganin, Evgeniya Ustinova, Hana Ajakan, Pascal Germain, Hugo
  Larochelle, Fran{\c{c}}ois Laviolette, Mario Marchand, and Victor~S.
  Lempitsky. 2016.
\newblock \href {http://jmlr.org/papers/v17/15-239.html} {Domain-adversarial
  training of neural networks}.
\newblock \emph{The Journal of Machine Learning Research}, 17:59:1--59:35.

\bibitem[{Gao et~al.(2021)Gao, Fisch, and Chen}]{DBLP:conf/acl/GaoFC20}
Tianyu Gao, Adam Fisch, and Danqi Chen. 2021.
\newblock \href {https://doi.org/10.18653/v1/2021.acl-long.295} {Making
  pre-trained language models better few-shot learners}.
\newblock In \emph{Proceedings of the 59th Annual Meeting of the Association
  for Computational Linguistics and the 11th International Joint Conference on
  Natural Language Processing, {ACL/IJCNLP} 2021, (Volume 1: Long Papers),
  Virtual Event, August 1-6, 2021}, pages 3816--3830. Association for
  Computational Linguistics.

\bibitem[{Glorot et~al.(2011)Glorot, Bordes, and
  Bengio}]{DBLP:conf/icml/GlorotBB11}
Xavier Glorot, Antoine Bordes, and Yoshua Bengio. 2011.
\newblock \href {https://icml.cc/2011/papers/342\_icmlpaper.pdf} {Domain
  adaptation for large-scale sentiment classification: {A} deep learning
  approach}.
\newblock In \emph{Proceedings of the 28th International Conference on Machine
  Learning, {ICML} 2011, Bellevue, Washington, USA, June 28 - July 2, 2011},
  pages 513--520. Omnipress.

\bibitem[{Gong et~al.(2020)Gong, Yu, and Xia}]{DBLP:conf/emnlp/GongYX20}
Chenggong Gong, Jianfei Yu, and Rui Xia. 2020.
\newblock \href {https://doi.org/10.18653/v1/2020.emnlp-main.572} {Unified
  feature and instance based domain adaptation for aspect-based sentiment
  analysis}.
\newblock In \emph{Proceedings of the 2020 Conference on Empirical Methods in
  Natural Language Processing, {EMNLP} 2020, Online, November 16-20, 2020},
  pages 7035--7045. Association for Computational Linguistics.

\bibitem[{Guo et~al.(2018)Guo, Shah, and Barzilay}]{DBLP:conf/emnlp/GuoSB18}
Jiang Guo, Darsh~J. Shah, and Regina Barzilay. 2018.
\newblock \href {https://www.aclweb.org/anthology/D18-1498/} {Multi-source
  domain adaptation with mixture of experts}.
\newblock In \emph{Proceedings of the 2018 Conference on Empirical Methods in
  Natural Language Processing, Brussels, Belgium, October 31 - November 4,
  2018}, pages 4694--4703. Association for Computational Linguistics.

\bibitem[{Han and Eisenstein(2019)}]{DBLP:conf/emnlp/HanE19}
Xiaochuang Han and Jacob Eisenstein. 2019.
\newblock \href {https://doi.org/10.18653/v1/D19-1433} {Unsupervised domain
  adaptation of contextualized embeddings for sequence labeling}.
\newblock In \emph{Proceedings of the 2019 Conference on Empirical Methods in
  Natural Language Processing and the 9th International Joint Conference on
  Natural Language Processing, {EMNLP-IJCNLP} 2019, Hong Kong, China, November
  3-7, 2019}, pages 4237--4247. Association for Computational Linguistics.

\bibitem[{Han et~al.(2021)Han, Zhao, Ding, Liu, and
  Sun}]{DBLP:journals/corr/abs-2105-11259}
Xu~Han, Weilin Zhao, Ning Ding, Zhiyuan Liu, and Maosong Sun. 2021.
\newblock \href {http://arxiv.org/abs/2105.11259} {{PTR:} prompt tuning with
  rules for text classification}.
\newblock \emph{CoRR}, abs/2105.11259.

\bibitem[{Haviv et~al.(2021)Haviv, Berant, and
  Globerson}]{DBLP:conf/eacl/HavivBG21}
Adi Haviv, Jonathan Berant, and Amir Globerson. 2021.
\newblock \href {https://aclanthology.org/2021.eacl-main.316/} {Bertese:
  Learning to speak to {BERT}}.
\newblock In \emph{Proceedings of the 16th Conference of the European Chapter
  of the Association for Computational Linguistics: Main Volume, {EACL} 2021,
  Online, April 19 - 23, 2021}, pages 3618--3623. Association for Computational
  Linguistics.

\bibitem[{Hu and Liu(2004)}]{DBLP:conf/kdd/HuL04}
Minqing Hu and Bing Liu. 2004.
\newblock \href {https://doi.org/10.1145/1014052.1014073} {Mining and
  summarizing customer reviews}.
\newblock In \emph{Proceedings of the 10th {ACM} {SIGKDD} International
  Conference on Knowledge Discovery and Data Mining, Seattle, Washington, USA,
  August 22-25, 2004}, pages 168--177. {ACM}.

\bibitem[{Hu et~al.(2018)Hu, Niu, Sato, and Sugiyama}]{DBLP:conf/icml/HuNSS18}
Weihua Hu, Gang Niu, Issei Sato, and Masashi Sugiyama. 2018.
\newblock \href {http://proceedings.mlr.press/v80/hu18a.html} {Does
  distributionally robust supervised learning give robust classifiers?}
\newblock In \emph{Proceedings of the 35th International Conference on Machine
  Learning, {ICML} 2018, Stockholmsm{\"{a}}ssan, Stockholm, Sweden, July 10-15,
  2018}, volume~80 of \emph{Proceedings of Machine Learning Research}, pages
  2034--2042. {PMLR}.

\bibitem[{Jiang et~al.(2020)Jiang, Xu, Araki, and
  Neubig}]{DBLP:journals/tacl/JiangXAN20}
Zhengbao Jiang, Frank~F. Xu, Jun Araki, and Graham Neubig. 2020.
\newblock \href {https://transacl.org/ojs/index.php/tacl/article/view/1983}
  {How can we know what language models know}.
\newblock \emph{Transactions of the Association for Computational Linguistics},
  8:423--438.

\bibitem[{Kim et~al.(2017)Kim, Stratos, and Kim}]{DBLP:conf/acl/KimSK17}
Young{-}Bum Kim, Karl Stratos, and Dongchan Kim. 2017.
\newblock \href {https://doi.org/10.18653/v1/P17-1060} {Domain attention with
  an ensemble of experts}.
\newblock In \emph{Proceedings of the 55th Annual Meeting of the Association
  for Computational Linguistics, {ACL} 2017, Vancouver, Canada, July 30 -
  August 4, Volume 1: Long Papers}, pages 643--653. Association for
  Computational Linguistics.

\bibitem[{Kingma and Ba(2015)}]{DBLP:journals/corr/KingmaB14}
Diederik~P. Kingma and Jimmy Ba. 2015.
\newblock \href {http://arxiv.org/abs/1412.6980} {Adam: {A} method for
  stochastic optimization}.
\newblock In \emph{3rd International Conference on Learning Representations,
  {ICLR} 2015, San Diego, CA, USA, May 7-9, 2015, Conference Track
  Proceedings}.

\bibitem[{Koh et~al.(2020)Koh, Sagawa, Marklund, Xie, Zhang, Balsubramani, Hu,
  Yasunaga, Phillips, Beery, Leskovec, Kundaje, Pierson, Levine, Finn, and
  Liang}]{DBLP:journals/corr/abs-2012-07421}
Pang~Wei Koh, Shiori Sagawa, Henrik Marklund, Sang~Michael Xie, Marvin Zhang,
  Akshay Balsubramani, Weihua Hu, Michihiro Yasunaga, Richard~Lanas Phillips,
  Sara Beery, Jure Leskovec, Anshul Kundaje, Emma Pierson, Sergey Levine,
  Chelsea Finn, and Percy Liang. 2020.
\newblock \href {http://arxiv.org/abs/2012.07421} {{WILDS:} {A} benchmark of
  in-the-wild distribution shifts}.
\newblock \emph{CoRR}, abs/2012.07421.

\bibitem[{Lekhtman et~al.(2021)Lekhtman, Ziser, and
  Reichart}]{lekhtman2021dilbert}
Entony Lekhtman, Yftah Ziser, and Roi Reichart. 2021.
\newblock \href {https://aclanthology.org/2021.emnlp-main.20} {{DILBERT:}
  customized pre-training for domain adaptation with category shift, with an
  application to aspect extraction}.
\newblock In \emph{Proceedings of the 2021 Conference on Empirical Methods in
  Natural Language Processing, {EMNLP} 2021, Virtual Event / Punta Cana,
  Dominican Republic, 7-11 November, 2021}, pages 219--230. Association for
  Computational Linguistics.

\bibitem[{Lester et~al.(2021)Lester, Al{-}Rfou, and
  Constant}]{DBLP:journals/corr/abs-2104-08691}
Brian Lester, Rami Al{-}Rfou, and Noah Constant. 2021.
\newblock \href {http://arxiv.org/abs/2104.08691} {The power of scale for
  parameter-efficient prompt tuning}.
\newblock \emph{CoRR}, abs/2104.08691.

\bibitem[{Lewis et~al.(2020)Lewis, Liu, Goyal, Ghazvininejad, Mohamed, Levy,
  Stoyanov, and Zettlemoyer}]{lewis-etal-2020-bart}
Mike Lewis, Yinhan Liu, Naman Goyal, Marjan Ghazvininejad, Abdelrahman Mohamed,
  Omer Levy, Veselin Stoyanov, and Luke Zettlemoyer. 2020.
\newblock \href {https://doi.org/10.18653/v1/2020.acl-main.703} {{BART:}
  denoising sequence-to-sequence pre-training for natural language generation,
  translation, and comprehension}.
\newblock In \emph{Proceedings of the 58th Annual Meeting of the Association
  for Computational Linguistics, {ACL} 2020, Online, July 5-10, 2020}, pages
  7871--7880. Association for Computational Linguistics.

\bibitem[{Li and Zong(2008)}]{DBLP:conf/acl/LiZ08}
Shoushan Li and Chengqing Zong. 2008.
\newblock \href {https://www.aclweb.org/anthology/P08-2065/} {Multi-domain
  sentiment classification}.
\newblock In \emph{{ACL} 2008, Proceedings of the 46th Annual Meeting of the
  Association for Computational Linguistics, June 15-20, 2008, Columbus, Ohio,
  USA, Short Papers}, pages 257--260. The Association for Computer Linguistics.

\bibitem[{Li and Liang(2021)}]{DBLP:conf/acl/LiL20}
Xiang~Lisa Li and Percy Liang. 2021.
\newblock \href {https://doi.org/10.18653/v1/2021.acl-long.353} {Prefix-tuning:
  Optimizing continuous prompts for generation}.
\newblock In \emph{Proceedings of the 59th Annual Meeting of the Association
  for Computational Linguistics and the 11th International Joint Conference on
  Natural Language Processing, {ACL/IJCNLP} 2021, (Volume 1: Long Papers),
  Virtual Event, August 1-6, 2021}, pages 4582--4597. Association for
  Computational Linguistics.

\bibitem[{Lin(2004)}]{Lin2004ROUGEAP}
Chin-Yew Lin. 2004.
\newblock \href {https://aclanthology.org/W04-1013.pdf} {Rouge: A package for
  automatic evaluation of summaries}.
\newblock In \emph{Proceedings of Workshop on Text Summarization Branches Out,
  Post Conference Workshop of ACL 2004}.

\bibitem[{Liu et~al.(2021{\natexlab{a}})Liu, Yuan, Fu, Jiang, Hayashi, and
  Neubig}]{DBLP:journals/corr/abs-2107-13586}
Pengfei Liu, Weizhe Yuan, Jinlan Fu, Zhengbao Jiang, Hiroaki Hayashi, and
  Graham Neubig. 2021{\natexlab{a}}.
\newblock \href {http://arxiv.org/abs/2107.13586} {Pre-train, prompt, and
  predict: {A} systematic survey of prompting methods in natural language
  processing}.
\newblock \emph{CoRR}, abs/2107.13586.

\bibitem[{Liu et~al.(2021{\natexlab{b}})Liu, Zheng, Du, Ding, Qian, Yang, and
  Tang}]{DBLP:journals/corr/abs-2103-10385}
Xiao Liu, Yanan Zheng, Zhengxiao Du, Ming Ding, Yujie Qian, Zhilin Yang, and
  Jie Tang. 2021{\natexlab{b}}.
\newblock \href {http://arxiv.org/abs/2103.10385} {{GPT} understands, too}.
\newblock \emph{CoRR}, abs/2103.10385.

\bibitem[{Luo et~al.(2008)Luo, Zhuang, Xiong, Xiong, and
  He}]{DBLP:conf/cikm/LuoZHXH08}
Ping Luo, Fuzhen Zhuang, Hui Xiong, Yuhong Xiong, and Qing He. 2008.
\newblock \href {https://doi.org/10.1145/1458082.1458099} {Transfer learning
  from multiple source domains via consensus regularization}.
\newblock In \emph{Proceedings of the 17th {ACM} Conference on Information and
  Knowledge Management, {CIKM} 2008, Napa Valley, California, USA, October
  26-30, 2008}, pages 103--112. {ACM}.

\bibitem[{McClosky et~al.(2010)McClosky, Charniak, and
  Johnson}]{mcclosky2010automatic}
David McClosky, Eugene Charniak, and Mark Johnson. 2010.
\newblock \href {https://www.aclweb.org/anthology/N10-1004/} {Automatic domain
  adaptation for parsing}.
\newblock In \emph{Human Language Technologies: Conference of the North
  American Chapter of the Association of Computational Linguistics,
  Proceedings, June 2-4, 2010, Los Angeles, California, {USA}}, pages 28--36.
  The Association for Computational Linguistics.

\bibitem[{Muandet et~al.(2013)Muandet, Balduzzi, and
  Sch{\"{o}}lkopf}]{DBLP:conf/icml/MuandetBS13}
Krikamol Muandet, David Balduzzi, and Bernhard Sch{\"{o}}lkopf. 2013.
\newblock \href {http://proceedings.mlr.press/v28/muandet13.html} {Domain
  generalization via invariant feature representation}.
\newblock In \emph{Proceedings of the 30th International Conference on Machine
  Learning, {ICML} 2013, Atlanta, GA, USA, 16-21 June 2013}, volume~28 of
  \emph{{JMLR} Workshop and Conference Proceedings}, pages 10--18. JMLR.org.

\bibitem[{M{\"{u}}ller et~al.(2020)M{\"{u}}ller, Rios, and
  Sennrich}]{DBLP:conf/amta/MullerRS20}
Mathias M{\"{u}}ller, Annette Rios, and Rico Sennrich. 2020.
\newblock \href {https://aclanthology.org/2020.amta-research.14/} {Domain
  robustness in neural machine translation}.
\newblock In \emph{Proceedings of the 14th Conference of the Association for
  Machine Translation in the Americas, {AMTA} 2020, Virtual, October 6-9,
  2020}, pages 151--164. Association for Machine Translation in the Americas.

\bibitem[{Oren et~al.(2019)Oren, Sagawa, Hashimoto, and
  Liang}]{DBLP:conf/emnlp/OrenSHL19}
Yonatan Oren, Shiori Sagawa, Tatsunori~B. Hashimoto, and Percy Liang. 2019.
\newblock \href {https://doi.org/10.18653/v1/D19-1432} {Distributionally robust
  language modeling}.
\newblock In \emph{Proceedings of the 2019 Conference on Empirical Methods in
  Natural Language Processing and the 9th International Joint Conference on
  Natural Language Processing, {EMNLP-IJCNLP} 2019, Hong Kong, China, November
  3-7, 2019}, pages 4226--4236. Association for Computational Linguistics.

\bibitem[{von Oswald et~al.(2020)von Oswald, Henning, Sacramento, and
  Grewe}]{DBLP:conf/iclr/OswaldHSG20}
Johannes von Oswald, Christian Henning, Jo{\~{a}}o Sacramento, and Benjamin~F.
  Grewe. 2020.
\newblock \href {https://openreview.net/forum?id=SJgwNerKvB} {Continual
  learning with hypernetworks}.
\newblock In \emph{8th International Conference on Learning Representations,
  {ICLR} 2020, Addis Ababa, Ethiopia, April 26-30, 2020}. OpenReview.net.

\bibitem[{Palatucci et~al.(2009)Palatucci, Pomerleau, Hinton, and
  Mitchell}]{DBLP:conf/nips/PalatucciPHM09}
Mark Palatucci, Dean Pomerleau, Geoffrey~E. Hinton, and Tom~M. Mitchell. 2009.
\newblock \href
  {https://proceedings.neurips.cc/paper/2009/hash/1543843a4723ed2ab08e18053ae6dc5b-Abstract.html}
  {Zero-shot learning with semantic output codes}.
\newblock In \emph{Advances in Neural Information Processing Systems 22: 23rd
  Annual Conference on Neural Information Processing Systems 2009. Proceedings
  of a meeting held 7-10 December 2009, Vancouver, British Columbia, Canada},
  pages 1410--1418. Curran Associates, Inc.

\bibitem[{Pan et~al.(2010)Pan, Ni, Sun, Yang, and
  Chen}]{DBLP:conf/www/PanNSYC10}
Sinno~Jialin Pan, Xiaochuan Ni, Jian{-}Tao Sun, Qiang Yang, and Zheng Chen.
  2010.
\newblock \href {https://doi.org/10.1145/1772690.1772767} {Cross-domain
  sentiment classification via spectral feature alignment}.
\newblock In \emph{Proceedings of the 19th International Conference on World
  Wide Web, {WWW} 2010, Raleigh, North Carolina, USA, April 26-30, 2010}, pages
  751--760. {ACM}.

\bibitem[{Peng et~al.(2018)Peng, Wu, and Ernst}]{DBLP:conf/eccv/PengWE18}
Kuan{-}Chuan Peng, Ziyan Wu, and Jan Ernst. 2018.
\newblock \href {https://doi.org/10.1007/978-3-030-01252-6\_47} {Zero-shot deep
  domain adaptation}.
\newblock In \emph{Computer Vision - {ECCV} 2018 - 15th European Conference,
  Munich, Germany, September 8-14, 2018, Proceedings, Part {XI}}, volume 11215
  of \emph{Lecture Notes in Computer Science}, pages 793--810. Springer.

\bibitem[{Pfeiffer et~al.(2020)Pfeiffer, Vulic, Gurevych, and
  Ruder}]{DBLP:conf/emnlp/PfeifferVGR20}
Jonas Pfeiffer, Ivan Vulic, Iryna Gurevych, and Sebastian Ruder. 2020.
\newblock \href {https://doi.org/10.18653/v1/2020.emnlp-main.617} {{MAD-X:} an
  adapter-based framework for multi-task cross-lingual transfer}.
\newblock In \emph{Proceedings of the 2020 Conference on Empirical Methods in
  Natural Language Processing, {EMNLP} 2020, Online, November 16-20, 2020},
  pages 7654--7673. Association for Computational Linguistics.

\bibitem[{Pontiki et~al.(2014)Pontiki, Galanis, Pavlopoulos, Papageorgiou,
  Androutsopoulos, and Manandhar}]{DBLP:conf/semeval/PontikiGPPAM14}
Maria Pontiki, Dimitris Galanis, John Pavlopoulos, Harris Papageorgiou, Ion
  Androutsopoulos, and Suresh Manandhar. 2014.
\newblock \href {https://doi.org/10.3115/v1/s14-2004} {Semeval-2014 task 4:
  Aspect based sentiment analysis}.
\newblock In \emph{Proceedings of the 8th International Workshop on Semantic
  Evaluation, SemEval@COLING 2014, Dublin, Ireland, August 23-24, 2014}, pages
  27--35. The Association for Computer Linguistics.

\bibitem[{Raffel et~al.(2020)Raffel, Shazeer, Roberts, Lee, Narang, Matena,
  Zhou, Li, and Liu}]{raffel2020exploring}
Colin Raffel, Noam Shazeer, Adam Roberts, Katherine Lee, Sharan Narang, Michael
  Matena, Yanqi Zhou, Wei Li, and Peter~J. Liu. 2020.
\newblock \href {https://www.jmlr.org/papers/volume21/20-074/20-074.pdf}
  {Exploring the limits of transfer learning with a unified text-to-text
  transformer}.
\newblock \emph{Journal of Machine Learning Research}, 21:1--67.

\bibitem[{Ramponi and Plank(2020)}]{DBLP:conf/coling/RamponiP20}
Alan Ramponi and Barbara Plank. 2020.
\newblock \href {https://doi.org/10.18653/v1/2020.coling-main.603} {Neural
  unsupervised domain adaptation in {NLP} - {A} survey}.
\newblock In \emph{Proceedings of the 28th International Conference on
  Computational Linguistics, {COLING} 2020, Barcelona, Spain (Online), December
  8-13, 2020}, pages 6838--6855. International Committee on Computational
  Linguistics.

\bibitem[{Reichart and Rappoport(2007)}]{reichart2007self}
Roi Reichart and Ari Rappoport. 2007.
\newblock \href {https://aclanthology.org/P07-1078/} {Self-training for
  enhancement and domain adaptation of statistical parsers trained on small
  datasets}.
\newblock In \emph{{ACL} 2007, Proceedings of the 45th Annual Meeting of the
  Association for Computational Linguistics, June 23-30, 2007, Prague, Czech
  Republic}. The Association for Computational Linguistics.

\bibitem[{Ring(1995)}]{DBLP:phd/dnb/Ring95}
Mark~B. Ring. 1995.
\newblock \href {https://d-nb.info/945690320} {\emph{Continual learning in
  reinforcement environments}}.
\newblock Ph.D. thesis, University of Texas at Austin, TX, {USA}.

\bibitem[{Roark and Bacchiani(2003)}]{roark2003supervised}
Brian Roark and Michiel Bacchiani. 2003.
\newblock \href {https://www.aclweb.org/anthology/N03-1027/} {Supervised and
  unsupervised {PCFG} adaptation to novel domains}.
\newblock In \emph{Human Language Technology Conference of the North American
  Chapter of the Association for Computational Linguistics, {HLT-NAACL} 2003,
  Edmonton, Canada, May 27 - June 1, 2003}. The Association for Computational
  Linguistics.

\bibitem[{R{\"{u}}ckl{\'{e}} et~al.(2020)R{\"{u}}ckl{\'{e}}, Pfeiffer, and
  Gurevych}]{DBLP:conf/emnlp/RucklePG20}
Andreas R{\"{u}}ckl{\'{e}}, Jonas Pfeiffer, and Iryna Gurevych. 2020.
\newblock \href {https://doi.org/10.18653/v1/2020.emnlp-main.194} {Multicqa:
  Zero-shot transfer of self-supervised text matching models on a massive
  scale}.
\newblock In \emph{Proceedings of the 2020 Conference on Empirical Methods in
  Natural Language Processing, {EMNLP} 2020, Online, November 16-20, 2020},
  pages 2471--2486. Association for Computational Linguistics.

\bibitem[{Rush et~al.(2012)Rush, Reichart, Collins, and
  Globerson}]{rush2012improved}
Alexander~M. Rush, Roi Reichart, Michael Collins, and Amir Globerson. 2012.
\newblock \href {https://www.aclweb.org/anthology/D12-1131/} {Improved parsing
  and {POS} tagging using inter-sentence consistency constraints}.
\newblock In \emph{Proceedings of the 2012 Joint Conference on Empirical
  Methods in Natural Language Processing and Computational Natural Language
  Learning, EMNLP-CoNLL 2012, July 12-14, 2012, Jeju Island, Korea}, pages
  1434--1444. {ACL}.

\bibitem[{Sagawa et~al.(2020)Sagawa, Koh, Hashimoto, and
  Liang}]{DBLP:conf/iclr/SagawaKHL20}
Shiori Sagawa, Pang~Wei Koh, Tatsunori~B. Hashimoto, and Percy Liang. 2020.
\newblock \href {https://openreview.net/forum?id=ryxGuJrFvS} {Distributionally
  robust neural networks}.
\newblock In \emph{8th International Conference on Learning Representations,
  {ICLR} 2020, Addis Ababa, Ethiopia, April 26-30, 2020}. OpenReview.net.

\bibitem[{Sanh et~al.(2019)Sanh, Debut, Chaumond, and
  Wolf}]{DBLP:journals/corr/abs-1910-01108}
Victor Sanh, Lysandre Debut, Julien Chaumond, and Thomas Wolf. 2019.
\newblock \href {http://arxiv.org/abs/1910.01108} {Distilbert, a distilled
  version of {BERT:} smaller, faster, cheaper and lighter}.
\newblock \emph{CoRR}, abs/1910.01108.

\bibitem[{Scao and Rush(2021)}]{DBLP:conf/naacl/ScaoR21}
Teven~Le Scao and Alexander~M. Rush. 2021.
\newblock \href {https://doi.org/10.18653/v1/2021.naacl-main.208} {How many
  data points is a prompt worth?}
\newblock In \emph{Proceedings of the 2021 Conference of the North American
  Chapter of the Association for Computational Linguistics: Human Language
  Technologies, {NAACL-HLT} 2021, Online, June 6-11, 2021}, pages 2627--2636.
  Association for Computational Linguistics.

\bibitem[{Schick and Sch{\"{u}}tze(2021)}]{DBLP:conf/eacl/SchickS21}
Timo Schick and Hinrich Sch{\"{u}}tze. 2021.
\newblock \href {https://aclanthology.org/2021.eacl-main.20/} {Exploiting
  cloze-questions for few-shot text classification and natural language
  inference}.
\newblock In \emph{Proceedings of the 16th Conference of the European Chapter
  of the Association for Computational Linguistics: Main Volume, {EACL} 2021,
  Online, April 19 - 23, 2021}, pages 255--269. Association for Computational
  Linguistics.

\bibitem[{Schnabel and Sch{\"{u}}tze(2014)}]{schnabel2014flors}
Tobias Schnabel and Hinrich Sch{\"{u}}tze. 2014.
\newblock \href
  {https://tacl2013.cs.columbia.edu/ojs/index.php/tacl/article/view/183}
  {{FLORS:} fast and simple domain adaptation for part-of-speech tagging}.
\newblock \emph{Transactions of the Association for Computational Linguistics},
  2:15--26.

\bibitem[{Shin et~al.(2020)Shin, Razeghi, IV, Wallace, and
  Singh}]{DBLP:conf/emnlp/ShinRLWS20}
Taylor Shin, Yasaman Razeghi, Robert L.~Logan IV, Eric Wallace, and Sameer
  Singh. 2020.
\newblock \href {https://doi.org/10.18653/v1/2020.emnlp-main.346} {Autoprompt:
  Eliciting knowledge from language models with automatically generated
  prompts}.
\newblock In \emph{Proceedings of the 2020 Conference on Empirical Methods in
  Natural Language Processing, {EMNLP} 2020, Online, November 16-20, 2020},
  pages 4222--4235. Association for Computational Linguistics.

\bibitem[{Sun and Lai(2020)}]{Sun2020ConditionedNL}
Fan{-}Keng Sun and Cheng{-}I Lai. 2020.
\newblock \href {http://arxiv.org/abs/2011.07347} {Conditioned natural language
  generation using only unconditioned language model: An exploration}.
\newblock \emph{CoRR}, abs/2011.07347.

\bibitem[{Toprak et~al.(2010)Toprak, Jakob, and
  Gurevych}]{DBLP:conf/acl/ToprakJG10}
Cigdem Toprak, Niklas Jakob, and Iryna Gurevych. 2010.
\newblock \href {https://www.aclweb.org/anthology/P10-1059/} {Sentence and
  expression level annotation of opinions in user-generated discourse}.
\newblock In \emph{{ACL} 2010, Proceedings of the 48th Annual Meeting of the
  Association for Computational Linguistics, July 11-16, 2010, Uppsala,
  Sweden}, pages 575--584. The Association for Computer Linguistics.

\bibitem[{Vaswani et~al.(2017)Vaswani, Shazeer, Parmar, Uszkoreit, Jones,
  Gomez, Kaiser, and Polosukhin}]{DBLP:conf/nips/VaswaniSPUJGKP17}
Ashish Vaswani, Noam Shazeer, Niki Parmar, Jakob Uszkoreit, Llion Jones,
  Aidan~N. Gomez, Lukasz Kaiser, and Illia Polosukhin. 2017.
\newblock \href
  {https://proceedings.neurips.cc/paper/2017/hash/3f5ee243547dee91fbd053c1c4a845aa-Abstract.html}
  {Attention is all you need}.
\newblock In \emph{Advances in Neural Information Processing Systems 30: Annual
  Conference on Neural Information Processing Systems 2017, December 4-9, 2017,
  Long Beach, CA, {USA}}, pages 5998--6008.

\bibitem[{Vijayakumar et~al.(2016)Vijayakumar, Cogswell, Selvaraju, Sun, Lee,
  Crandall, and Batra}]{DBLP:journals/corr/VijayakumarCSSL16}
Ashwin~K. Vijayakumar, Michael Cogswell, Ramprasaath~R. Selvaraju, Qing Sun,
  Stefan Lee, David~J. Crandall, and Dhruv Batra. 2016.
\newblock \href {http://arxiv.org/abs/1610.02424} {Diverse beam search:
  Decoding diverse solutions from neural sequence models}.
\newblock \emph{CoRR}, abs/1610.02424.

\bibitem[{Wald et~al.(2021)Wald, Feder, Greenfeld, and Shalit}]{wald2021on}
Yoav Wald, Amir Feder, Daniel Greenfeld, and Uri Shalit. 2021.
\newblock \href {https://openreview.net/forum?id=XWYJ25-yTRS} {On calibration
  and out-of-domain generalization}.
\newblock In \emph{Thirty-Fifth Conference on Neural Information Processing
  Systems}.

\bibitem[{Williams et~al.(2018)Williams, Nangia, and Bowman}]{N18-1101}
Adina Williams, Nikita Nangia, and Samuel Bowman. 2018.
\newblock \href {http://aclweb.org/anthology/N18-1101} {A broad-coverage
  challenge corpus for sentence understanding through inference}.
\newblock In \emph{Proceedings of the 2018 Conference of the North American
  Chapter of the Association for Computational Linguistics: Human Language
  Technologies, Volume 1 (Long Papers)}, pages 1112--1122. Association for
  Computational Linguistics.

\bibitem[{Wolf et~al.(2020)Wolf, Debut, Sanh, Chaumond, Delangue, Moi, Cistac,
  Rault, Louf, Funtowicz, Davison, Shleifer, von Platen, Ma, Jernite, Plu, Xu,
  Scao, Gugger, Drame, Lhoest, and Rush}]{wolf-etal-2020-transformers}
Thomas Wolf, Lysandre Debut, Victor Sanh, Julien Chaumond, Clement Delangue,
  Anthony Moi, Pierric Cistac, Tim Rault, Rémi Louf, Morgan Funtowicz, Joe
  Davison, Sam Shleifer, Patrick von Platen, Clara Ma, Yacine Jernite, Julien
  Plu, Canwen Xu, Teven~Le Scao, Sylvain Gugger, Mariama Drame, Quentin Lhoest,
  and Alexander~M. Rush. 2020.
\newblock \href {https://www.aclweb.org/anthology/2020.emnlp-demos.6}
  {Transformers: State-of-the-art natural language processing}.
\newblock In \emph{Proceedings of the 2020 Conference on Empirical Methods in
  Natural Language Processing: System Demonstrations}, pages 38--45, Online.
  Association for Computational Linguistics.

\bibitem[{Wright and Augenstein(2020)}]{wright2020transformer}
Dustin Wright and Isabelle Augenstein. 2020.
\newblock \href {https://www.aclweb.org/anthology/2020.emnlp-main.639/}
  {Transformer based multi-source domain adaptation}.
\newblock In \emph{Proceedings of the 2020 Conference on Empirical Methods in
  Natural Language Processing (EMNLP)}, pages 7963--7974.

\bibitem[{Yang and Eisenstein(2014)}]{DBLP:conf/acl/YangE14}
Yi~Yang and Jacob Eisenstein. 2014.
\newblock \href {https://doi.org/10.3115/v1/p14-2088} {Fast easy unsupervised
  domain adaptation with marginalized structured dropout}.
\newblock In \emph{Proceedings of the 52nd Annual Meeting of the Association
  for Computational Linguistics, {ACL} 2014, June 22-27, 2014, Baltimore, MD,
  USA, Volume 2: Short Papers}, pages 538--544. The Association for Computer
  Linguistics.

\bibitem[{Zhang et~al.(2020)Zhang, Kishore, Wu, Weinberger, and
  Artzi}]{DBLP:conf/iclr/ZhangKWWA20}
Tianyi Zhang, Varsha Kishore, Felix Wu, Kilian~Q. Weinberger, and Yoav Artzi.
  2020.
\newblock \href {https://openreview.net/forum?id=SkeHuCVFDr} {Bertscore:
  Evaluating text generation with {BERT}}.
\newblock In \emph{8th International Conference on Learning Representations,
  {ICLR} 2020, Addis Ababa, Ethiopia, April 26-30, 2020}. OpenReview.net.

\bibitem[{Zhao et~al.(2018)Zhao, Zhang, Wu, Costeira, Moura, and
  Gordon}]{DBLP:conf/iclr/0002ZWCMG18}
Han Zhao, Shanghang Zhang, Guanhang Wu, Jo{\~{a}}o~Paulo Costeira, Jos{\'{e}}
  M.~F. Moura, and Geoffrey~J. Gordon. 2018.
\newblock \href {https://openreview.net/forum?id=BkzXYSkvz} {Multiple source
  domain adaptation with adversarial learning}.
\newblock In \emph{6th International Conference on Learning Representations,
  {ICLR} 2018, Vancouver, BC, Canada, April 30 - May 3, 2018, Workshop Track
  Proceedings}. OpenReview.net.

\bibitem[{Ziser and Reichart(2017)}]{DBLP:conf/conll/ZiserR17}
Yftah Ziser and Roi Reichart. 2017.
\newblock \href {https://doi.org/10.18653/v1/K17-1040} {Neural structural
  correspondence learning for domain adaptation}.
\newblock In \emph{Proceedings of the 21st Conference on Computational Natural
  Language Learning (CoNLL 2017), Vancouver, Canada, August 3-4, 2017}, pages
  400--410. Association for Computational Linguistics.

\bibitem[{Ziser and Reichart(2018)}]{ziser2018pivot}
Yftah Ziser and Roi Reichart. 2018.
\newblock \href {https://doi.org/10.18653/v1/n18-1112} {Pivot based language
  modeling for improved neural domain adaptation}.
\newblock In \emph{Proceedings of the 2018 Conference of the North American
  Chapter of the Association for Computational Linguistics: Human Language
  Technologies, {NAACL-HLT} 2018, New Orleans, Louisiana, USA, June 1-6, 2018,
  Volume 1 (Long Papers)}, pages 1241--1251. Association for Computational
  Linguistics.

\bibitem[{Ziser and Reichart(2019)}]{DBLP:conf/acl/ZiserR19}
Yftah Ziser and Roi Reichart. 2019.
\newblock \href {https://doi.org/10.18653/v1/p19-1591} {Task refinement
  learning for improved accuracy and stability of unsupervised domain
  adaptation}.
\newblock In \emph{Proceedings of the 57th Conference of the Association for
  Computational Linguistics, {ACL} 2019, Florence, Italy, July 28- August 2,
  2019, Volume 1: Long Papers}, pages 5895--5906. Association for Computational
  Linguistics.

\bibitem[{Zubiaga et~al.(2017)Zubiaga, Liakata, and
  Procter}]{DBLP:conf/socinfo/ZubiagaLP17}
Arkaitz Zubiaga, Maria Liakata, and Rob Procter. 2017.
\newblock \href {https://doi.org/10.1007/978-3-319-67217-5\_8} {Exploiting
  context for rumour detection in social media}.
\newblock In \emph{Social Informatics - 9th International Conference, SocInfo
  2017, Oxford, UK, September 13-15, 2017, Proceedings, Part {I}}, volume 10539
  of \emph{Lecture Notes in Computer Science}, pages 109--123. Springer.

\bibitem[{Zubiaga et~al.(2016)Zubiaga, Liakata, Procter, Wong Sak~Hoi, and
  Tolmie}]{zubiaga2016analysing}
Arkaitz Zubiaga, Maria Liakata, Rob Procter, Geraldine Wong Sak~Hoi, and Peter
  Tolmie. 2016.
\newblock \href
  {https://journals.plos.org/plosone/article?id=10.1371/journal.pone.0150989}
  {Analysing how people orient to and spread rumours in social media by looking
  at conversational threads}.
\newblock \emph{PloS one}, 11(3):e0150989.

\end{thebibliography}
